# How Santa Fe Ants Evolve

Dominic Wilson, Devinder Kaur,

***Abstract*— The Santa Fe Ant model problem has been extensively used to investigate, test and evaluate evolutionary computing systems and methods over the past two decades. There is however no literature on its program structures that are systematically used for fitness improvement, the geometries of those structures and their dynamics during optimization. This paper analyzes the Santa Fe Ant Problem using a new phenotypic schema and landscape analysis based on executed instruction sequences. For the first time we detail systematic structural features that give high fitness and the evolutionary dynamics of such structures. The new schema avoids variances due to introns.**

**We develop a phenotypic variation method that tests the new understanding of the landscape. We also develop a modified function set that tests newly identified synchronization constraints. We obtain favorable computational efforts compared to those in the literature, on testing the new variation and function set on both the Santa Fe Trail, and the more computationally demanding Los Altos Trail. Our findings suggest that for the Santa Fe Ant problem, a perspective of program assembly from repetition of highly fit responses to trail conditions leads to better analysis and performance.**

## I. Introduction

Evolutionary metaheuristics use evolution inspired variation and selection techniques in assembling structures (i.e. combinations of symbols such as program primitives) to solve or optimize a problem. For these metaheuristics to have computational efficiency advantages they have to narrow down their evaluations to more promising areas of the combination space. The focus on more promising search areas is usually presumed to come from the combination of certain high fitness structures (known as schemas), and/or a quality of the neighborhood of high fitness structures, such that small changes to them can lead to structures of even higher fitness.

Demonstrating support for these schema combination and neighborhood search processes as sources of search improvement has proven particularly difficult for metaheuristics that produce programs as their output. Evolving the Santa Fe ant problem using Genetic Programming is a textbook case of how a seemingly simple problem can show complicated dynamics and not conform to traditional schema or neighborhood search paradigms.

The Santa Fe ant problem is a well-known model problem that have been studied over the past two decades and is still being actively researched [1,2,3,4,5,6,7,8]. This problem is known for its status of being "hard" by virtue of evolutionary computing methods not solving it at much higher efficiencies than random search [2]. The hardness has been attributed to a fitness landscape that is difficult to search, being "rugged with multiple plateaus split by deep valleys and many local and global optima". Fixed length schema analysis has found the structure of programs "highly deceptive" [2].

Other recent papers support this idea of a landscape that defies a clear description of its structure and the evolutionary dynamics on such a structure. Miller and Thompson supported the findings that "the fitness space associated with the Santa Fe Trail has a great deal of randomness associated with it" [9]. Galvan-Lopez *et al.* found low-locality in GP on using various forms of mutation on the landscape [6]. They characterize the landscape as being "multimodally-deceptive". McDermitt *et al.* found the mapping from an ant's behavioral phenotype to its concrete path was "inherently badly-behaved", and concluded "alternative genetic encodings and operators cannot make the problem easy" [7].

The literature presents a picture of a landscape with little association between program structures and their fitness, and does not detail any specifics on what any metaheuristic uses for fitness improvement during evolutionary runs. We show in this paper that the relationship between program structures and their fitness is tractable; we detail systematic evolutionary dynamics that use certain structural properties for fitness improvement. We test our new understanding by developing a new variation method and a new set of operators that make the problem easier for genetic search.

In investigating this problem we developed a new set of analytic tools and visualizations that can possibly be of more general use upon further study. This includes using executed instruction sequences as schema. Additionally we contribute new understanding to the key role of implicit repetition in the evolutionary process, and the characteristics of local optima when solutions are not realized.

### A. Structure of this paper

Section II introduces the Santa Fe ant problem and discusses the literature on it. In Section III we discuss the representation we will be using. We also explain and provide examples of our new schema. We conduct experiments leading to the analysis and discussion of the relationship between program structure and fitness in section IV. This section will present how schemas are related to fitness and explain the regular neighborhood structure between schemas that represent fitness optima. This section will feature a new way of visualizing neighborhoods using Pascal's pyramids.

In section V we analyze the characteristics of solutions (which are the globally optimal programs that successfully consume all food on the trail). We show that most solutions have simple schema structures which we characterize. In Section VI we evolve solutions to the Ant problem using Genetic Programming. We keep track of the best program for each generation and analyzing how these programs change. We show the variations that change one high fitness program to another.

In Section VII we devise a search and a representation method that improve performance by using our findings on the neighborhood structure and synchronization requirements of solutions. We device and test a new phenotypic crossover method that explores high fitness areas. We discuss and test a representation that reduces synchronization requirement and leads to more solutions. We test both method on both the Santa Fe and more difficult Los Altos trail and discuss performance. Section VIII states our conclusions and future directions.

## II. THE SANTA FE ANT TRAIL LITERATURE

Fig. 1 is a diagram of the Santa Fe Trail. This trail is embedded in a toroidally connected grid of 32 x 32 cells. The food trail of 144 cells (89 containing food and 55 being gaps in the trail with no food). There are 10 left and 11 right turns of the trail. The objective of the Santa Fe ant problem is to evolve a program (i.e. a set of computer instructions) that can navigate this trail, finding all the food using a specified amount of energy.

Various amounts of energy have been used for this problem in the literature including 400 as in [4], 600 as in [2, 3, 9, 31, 11, 12, 13, 14, 17, 19, 20] and 615 as in [28, 29, 30, 33]. We will use the most common energy value of 600 in our experiments in this paper. The objective is to find a solution (i.e. a program that can find all food without exhausting its energy). Programs are usually personified as ants for this problem. The ant starts at the top left corner facing east. An ant can turn left or right or move forward one step in the direction it is facing. The ant can also sense whether there is food ahead in the direction it is facing. Each activity (other than sensing food) reduces the total energy of the ant by a unit amount.

### A. Related Literature

Jefferson et al. published the first paper on using evolutionary principles to develop trail following behavior [10]. They used genetic algorithms to train programs represented as finite state automata and artificial neural networks to solve the John Muir trail.

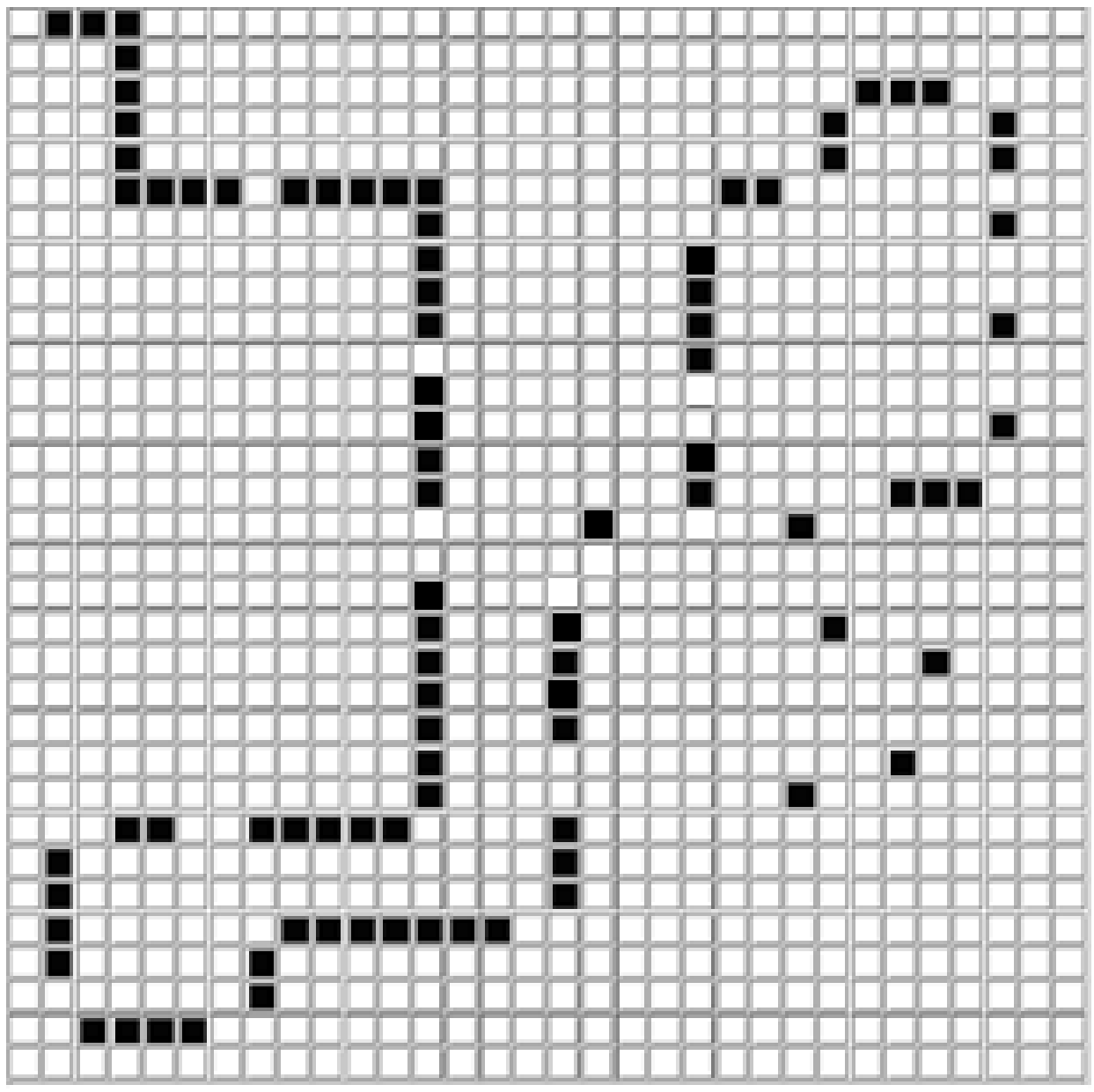

**Fig. 1: The Santa Fe ant trail**

The Santa Fe artificial ant trail is a well-studied Genetic Programming problem that was first used in Koza's seminal work [1]. This problem, along with the more computationally demanding Los Altos Trail problem, was also featured in Koza [8]. In this paper we will be using both problems along with the settings in Koza [8] for benchmarking.

The Santa Fe ant problem is the single most used problem for investigating bloat [3,11,12,13,14,15]. Luke and Panait [16] used this problem as one among several to compare bloat control methods.

The Santa Fe ant problem is sometimes used as a representative problem, usually among a suite of other problems. Common cases for this use are as a performance benchmark and to show the viability of metaheuristics or new methods. Chellapilla [17] used the problem to demonstrate that tree mutations can be effective in GP. Fonlupt and Robilliard used the problem as part of a suite to demonstrate the viability of Linear Differential Evolutionary Programming [18]. The problem has similarly been used to show the viability of Analytic Programming [19,20], Cartesian Genetic Programming [9] Grammatical Evolution [21] and Constituent Grammatical Evolution [22]. The problem has been used to study generalization ability [4] and negative slope coefficient [5].

The Santa Fe artificial ant trail has been well studied using other metaheuristics as well. Analyses in Grammatical Evolution have used the problem to investigate the effects of crossover types (and to a limited extent mutation rates) [23,24,25,26,27], degeneracy [27,28] and wrapping [23,27] on proportion of invalid individuals, genome length, genotypic variety and cumulative frequency of success. In [29] the effect of changes to the standard grammar and bias by the addition of grammar defined introns is investigated with and without degeneracy. The effect of grammar size and complexity on performance is investigated in [30].

*1) The Program Schema and Landscape*

Langdon and Poli [2] published a detailed analysis of ant problem, stating reasons why it was computationally difficult to solve efficiently, with simulated annealing, hill climbing and various GP methods not performing much better than random search. Their reasons included a large and rugged fitness landscape with many local and global optima, the highly deceptive nature of the search space on fixed length schema analysis and the lack of assembly of solutions using building blocks of above average fitness. Langdon [31] redefined the problem so the ant is obliged to traverse the trail in approximately the correct order. They showed improved performance on using a simple genetic programming system, with no size or depth restriction.

Langdon and Poli [2] found that the average fitness of programs increases with their size. In Section IV we will discuss why this is the case. In Section VI we will show how our findings explain the fact (stated in Langdon [11]) that programs with zero and very low fitness are produced even in the final generation of evolutionary runs on this problem.

The analysis of landscape structure, locality, and search on the ant problem has not been restricted to GP. Rothlauf and Oetzel [32] examined the locality properties of the binary representation used in GE. They found that a genotypic bit-mutation operator produced non-local changes in the phenotype. They proposed that further research be done to find other representations and mutation operators which would produce higher locality.

Hugosson *et al.* [33] also addressed GE locality by investigating the impact of four mutation operators using binary and Gray code representation, on performance. The combinations of representations and mutation operators were tested using three benchmark problems, including the Santa Fe Ant problem. The alternative representations did not result in higher locality or better performance. They raised the question whether higher locality in GE improves performance.

## III. Representation and Schema

In this section we discuss the representation and phenotypic schema we will be using in this paper.

The most common representation for ant programs in the literature (and the representation we will be using) is a function tree composed of the set of terminal functions (Move, Left, Right) and the set of nonterminal functions (IfFoodAhead, Prog2, Prog3). This representation has been used with many metaheuristics other than GP, such as Cartesian Genetic Programming, simulated annealing, and hill climbing e.g. [2, 3,17]. With this representation, an ant moves forward by one cell, or turns to its left, or right on executing the move, left or right functions respectively.

The "IfFoodAhead" function senses whether there is food on the cell the ant is currently facing, and conditionally selects one of two arguments for execution. Prog2 and Prog3 are functions that take two and three arguments respectively, which are then executed sequentially.

Unlike conditional execution (done by IfFoodAhead) and sequential execution (done by Prog2 and Prog3), repetition is implicit with no specific function representation. Repetition is done by repeatedly executing the entire program tree until the ant's energy is exhausted, or there is no more food on the trail. While this implicit repetition feature has not been highlighted much in the literature, we will see in Sections V and VI that it is important for understanding the relationship between structure and fitness.

Fig 2 shows three examples of programs using the GP representation. On Fig 2, "r", "l" and "m" represent the instructions right, left and move respectively. "If" stands for IfFoodAhead and Prog2 and Prog3 are represented by unlabeled nodes with two and three branches respectively. The left branch of the "If" function is taken when there is food ahead and the right branch is taken when there is no food ahead. Programs are traversed in order from left to right.

The programs in Fig 2 are used throughout this paper as examples to illustrate various concepts.

### A. Executed Instruction Schema

Traditionally a schema is a subset of program structures with similarities at certain positions [1,2,34,35, 36,37,38,39,

40,41,42,43,44]. These subsets have been used to partition evolutionary search spaces for the purpose of understanding and modeling the dynamics of populations. Although traditional GP schemas have provided exact mathematical models and probabilistic descriptions of the operations of selection and variation operations in GP, they have not been able to incorporate fitness information into their functionality. Their role is consequently limited in explaining evolutionary dynamics when fitness drives evolutionary change.

Most GP schemata use similarities in the internal nonterminal structures of program trees e.g. [1,2, 45, 46,47,48, 49,50,51,52,53,54,55]. In contrast we use the sequence of executed instructions as the basis for identifying schema. The identification of executed instruction sequences can be automated by recording which instructions get executed, and their order of execution, as a program runs. In some cases it is possible to discover the sequence by visual inspection and deducing the flow of control of a program.

An example of an executed instruction sequence is given in Fig 3. Fig 3 is a single string showing the sequence of instructions that are executed when the program in Fig 2(a) is run. The sequence in Fig 3 is divided into sections by commas; this division corresponds to the instances of repetition of the program. The commas make it easy to identify sections of the sequence that are identical.

The executed instruction sequence is the sequence of responses of a program to its input or environment. We define executed instruction schema as sections of the executed instruction sequence that respond to particular input or environment conditions a program is subjected to. Executed instruction schema are by definition phenotypic, as they are the responses programs makes to input or environment conditions.

As an example of an executed sequence schema consider the sections with instructions rrlml in Fig 3. This schema is the response of the program in Fig 2(a) when there is no food in its immediate surrounding cells. For all programs in Fig 2(a), Fig 2(b), and Fig 2(c), the instructions that will be executed in a situation with no food are those instructions that are underlined. The program in Fig 2(b) has the same "no food" schema as that of Fig 2(a), as it also responds with rrlml. For Fig 2(c), the schema that responds to the "no food" condition is llrmr.

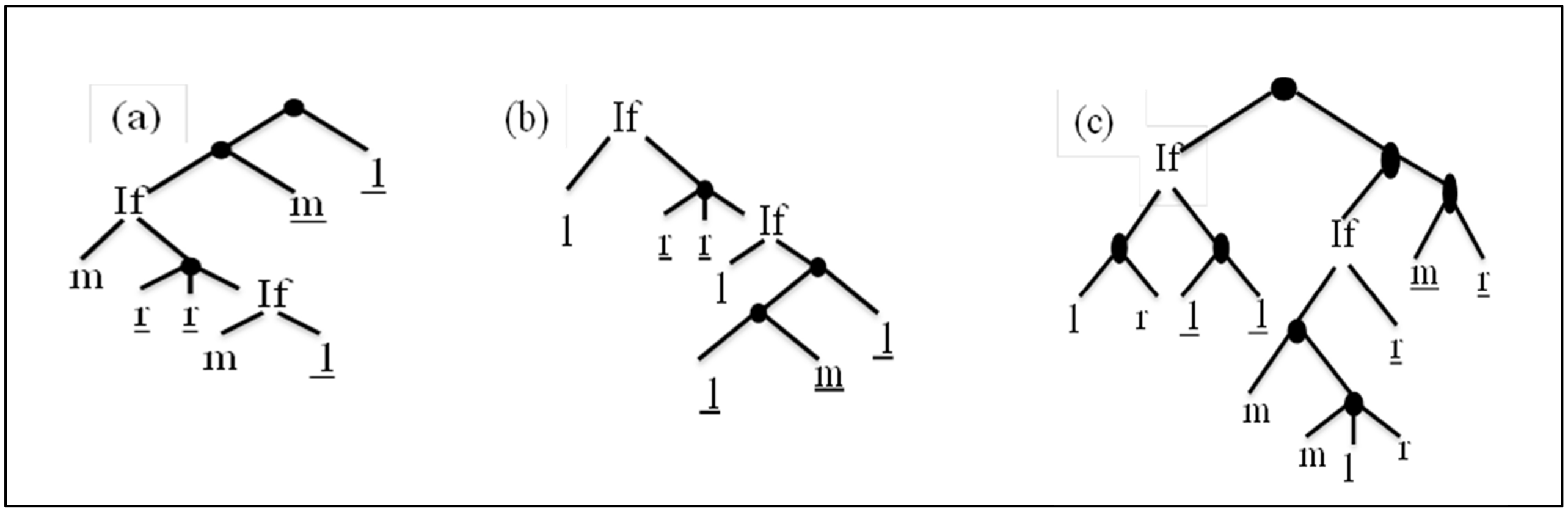

**Fig 2: Examples of Programs Using GP Representation**

We will refer to the schema that programs use when there is no food in their immediate surrounding cells as the ***default schema*** of the program, since more than 90% of the Santa Fe grid is empty. We will not need to give specific names to other schemas. The default schema of programs will be referred to often in this paper, and is necessary to understand the concepts explained.

```
mml,rrlml,rrmml,rrlml,rrlml,rrlml,mml,rrlml,rrlml,rrlml,rrlml,rrlml,rrlml,rrlml,rrmml,rrlml,rrlml,rrlml
,rrlml,rrlml,rrlml,rrlml,rrlml,rrlml,rrlml,rrlml,rrlml,rrlml,rrlml,rrlml,rrlml,rrlml,rrmml,rrlml,rrlml,
rrlml,rrlml,rrlml,rrlml,rrlml,rrlml,rrlml,mml,rrlml,rrlml,rrlml,rrlml,mml,rrlml,rrlml,rrlml,rrlml,mml,r
rlml,rrmml,rrlml,rrlml,rrlml,rrlml,rrlml,rrlml,rrlml,mml,rrlml,rrlml,rrlml,rrlml,rrlml,rrlml,rrlml,rrlm
l,rrlml,rrlml,rrmml,rrlml,rrlml,mml,rrlml,rrlml,rrlml,rrlml,rrlml,rrlml,rrlml,rrlml,rrmml,rrlml,rrlml,m
ml,rrlml,rrmml,rrlml,rrlml,rrlml,rrmml,rrlml,rrlml,rrlml,rrlml,rrlml,rrlml,rrlml,rrlml,rrlml,rrlml,rrmm
l,rrlml,rrlml,rrlml,rrlml,mml,rrlml,rrlml,mml,rrlml,rrlml,rrmml,rrlml,rrlml,rrmml,rrlml,rrlml,mml
```

**Fig 3: A single string showing the sequence of instructions executed when the program in Fig 2(a) is run.**

Programs without any "IfFoodAhead" decision nodes will respond the same way to all trail conditions, and therefore have the same schema for all conditions. In this case this common schema is made up of all the terminal nodes in the program.

If the program in Fig 2 (a) is directly facing a cell with food the sequence of instructions that are executed without any program repetition is mml. This is the schema for this particular condition. Similarly the schema that is used by Fig 2 (b) and (c) when either program is facing a cell with food is l and lrmmlrmr respectively. From these it can be seen that schemas can be of different lengths and can overlap with each other.

It should be noted that though internal nonterminal structures are not part of our schema they are important because they define (through the use of conditionals and sequential operators) which terminal nodes get included in the schema. Also note that because of the repetition of schema within the executed instruction sequence, a single change to a schema can result in several changes to the instruction sequence. The more a schema is used (and repeated in the instruction sequence) the more a change to it affects the sequence.

By using schemas that are (by definition) responses to trail conditions, we automatically eliminated any variance due to the presence of introns (i.e. non-executing instruction sequences) from subsequent analyses. Introns [56,57] (which are commonplace when programs are randomly created and evolved) are simply not part of any schema since they are never executed.

## IV. Program Fitness and Structure

Understanding the relationship between a program's fitness and its structure (i.e. the arrangements of its terminal and non-terminal instructions) is an important prerequisite to understanding evolutionary dynamics. In this section we will perform experiments and analyses that show how fitness is associated with structure. The experiments in this section will not use any variation or selection operation; as a result the findings in this section are general to the representation being used and not specific to GP or any particular metaheuristic.

### *A. Overview of approach*

To relate fitness to program structure we will use experiments and analyses to establish that:

a) Average fitness of randomly generated programs is related to their cell visiting behavior.
b) Cell visiting behavior is determined by schema size and composition.

We will characterize the cell visiting behavior of randomly generated programs as follows:

a) Programs with large default schemas approximate a random walk on the grid;
b) Programs with small default schemas use repetition to perform a non-random walk; the majority of these programs will cycle over a small set of the same cells leading to low fitness. Most high fitness programs have small default schemas and do not cycle over a small set of cells.

We point out the schema compositions that do (and do not) cycle over a small set of cells and detail the regular structure of their neighborhoods.

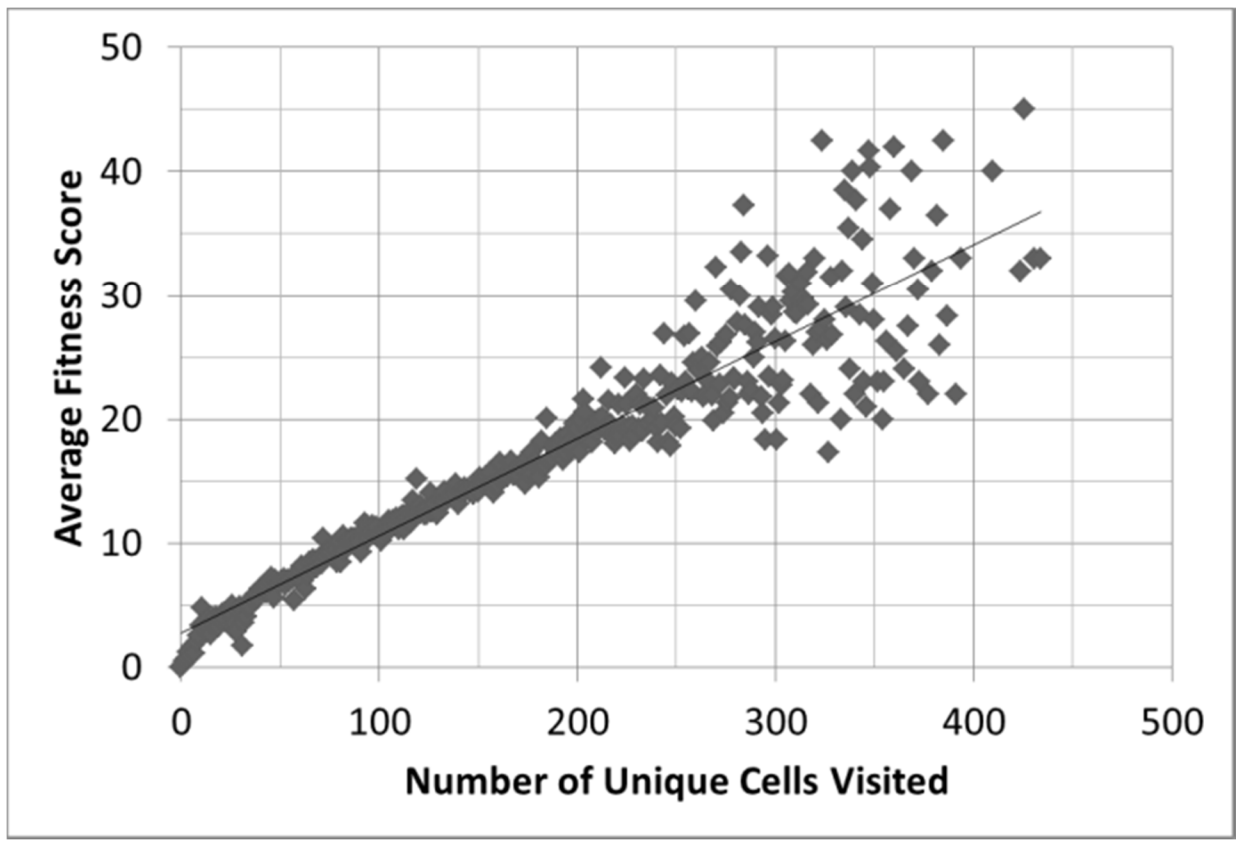

**Fig 4: Average fitness by number of cells visited by program**

### *B. Experiment 1*

Our first experiment (which we will refer to as Experiment 1) is to generate a random population and measure some properties of the individual programs. We generated a population of 90,000 random programs using the ramped half-and-half creation method with a depth ramp of 2 to 6. We measured the properties of each program, including its fitness score, the number of cells it visits, and how many times each program is repeated and its schema. Throughout this paper a higher fitness score means a fitter program.

#### *1) Relating Fitness to Cell Visits*

Our first analysis from Experiment 1 is to determine how the number of different cells visited by a program is related to its expected fitness. Fig 4 shows that the average fitness of randomly generated programs increases with the number of different cells the programs visit. This confirms the intuitive relation that the more cells a randomly generated program visits the higher its fitness score is likely to be.

This discovery leads us to consider that the key to understanding the relationship between program structure and fitness is in knowing how the structure of programs determine how many cells they visit. We examine this relation in the next two subsections.

#### *2) Cell Visits of Programs with Large Default Schemas*

Fig 5 shows the relationship between default schema size and the number of different cells visited. Fig 5 portrays that on average, programs with large default schemas sizes have a higher number of cell visits than programs with small default schemas. To understand why this is the case we will first consider how programs with large default schemas walk on the grid.

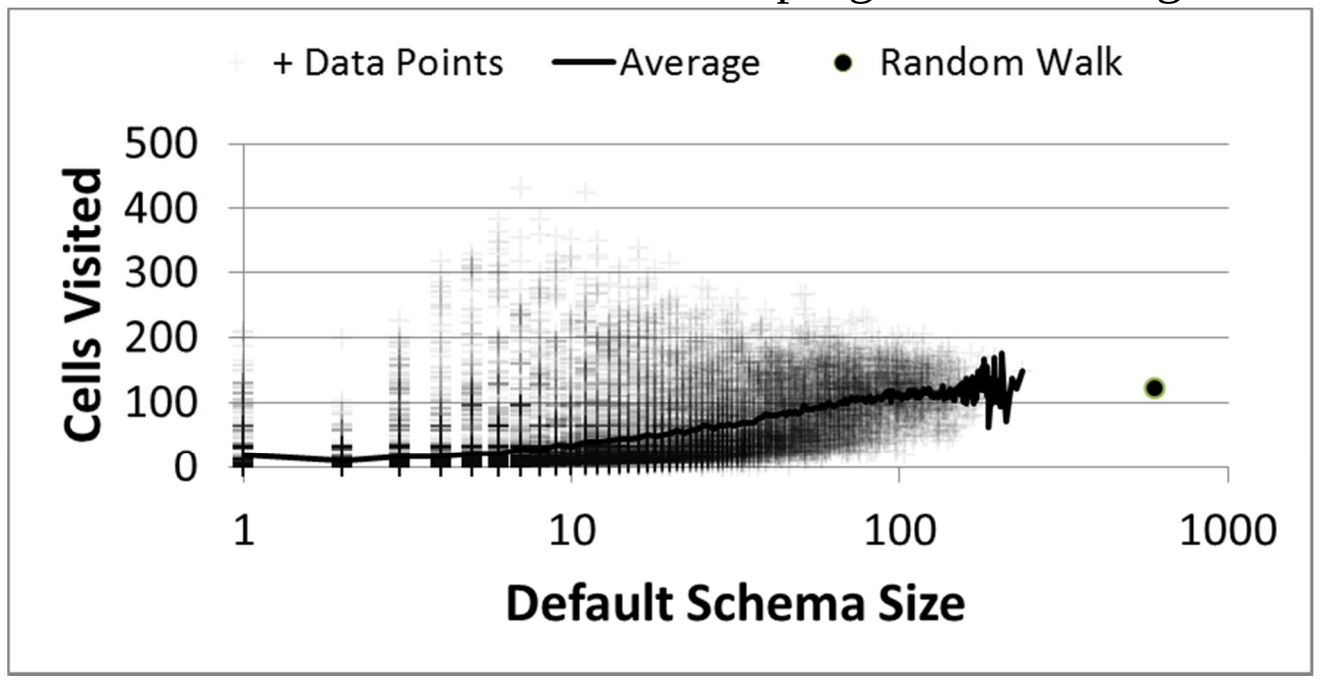

**Fig 5: Cells visited for various default schema sizes**

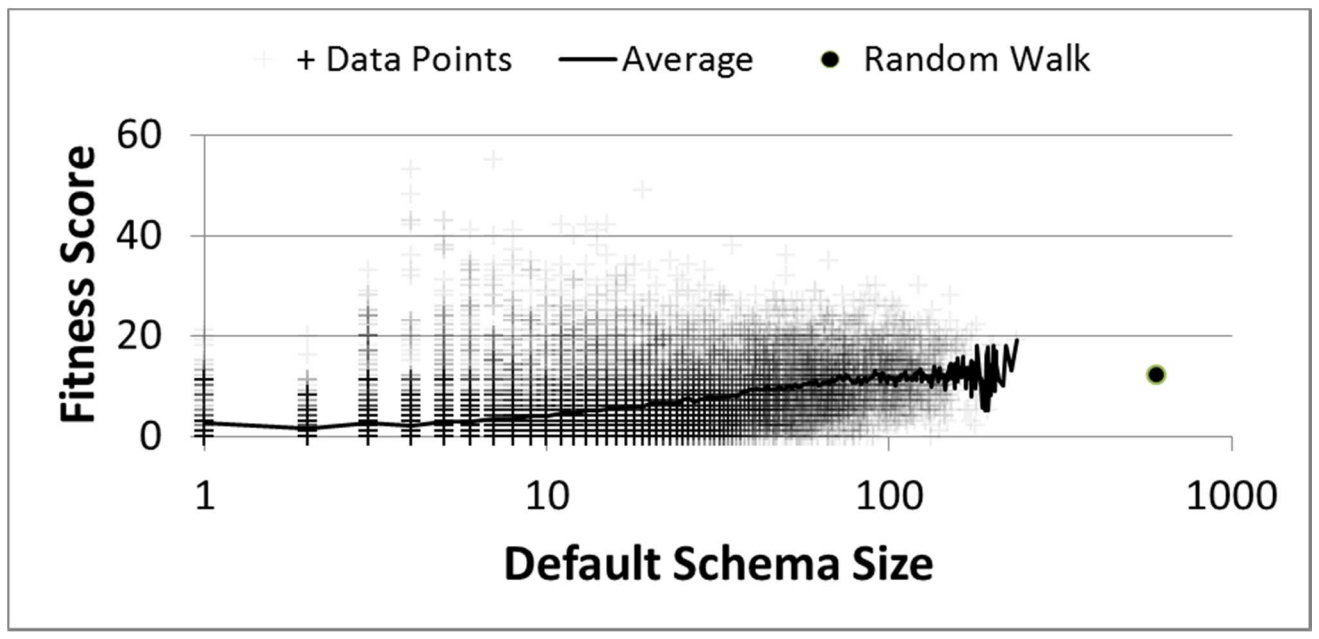

**Fig 6: Fitness scores for various default schema sizes**

Our hypothesis is that programs with large default schemas do a walk on the trail that is similar to a random walk. To confirm this hypothesis we conducted the following experiment (which will be referred to as Experiment 2). We performed 10,000 runs of a "random walk" on the trail†. For this walk we had a program start from the top-left cell of the grid pointing east. We then executed either a "move" or "left" or "right" instruction chosen at random and with equal probability. The random instruction selection and execution was done 600 times to simulate a nonrepeating program with an energy budget of 600.

Fig 5 shows a circle denoting the average number of cells visited during the random walk experiment. The 123.3 average cell visit value of the random walk fits our expectation that as a program's default schema gets larger its behavior approximates a random walk on the trail.

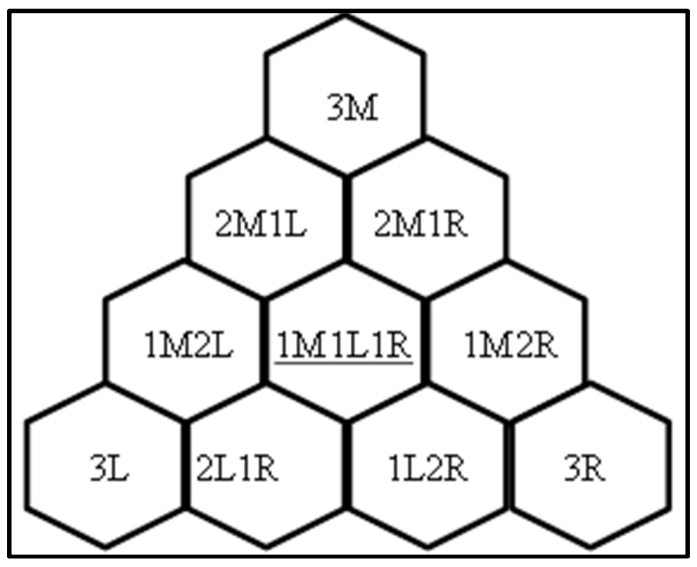

**Fig 7: A single Pascal's Pyramid layer showing the arrangement of Move (M), Left (L) and Right (R) instructions that give a schema size of 3.**

Fig 6 shows the relationship between default schema size and fitness. Fig 6 shows a circle denoting the average fitness obtained during the random walk experiment. We see that as a program's default schema gets larger its fitness approximates 12.3‡, the average fitness of a random walk on the trail. This further confirms our expectation that program with large default schema approximates a random walk on the trail. We examine the behavior of programs with small default schema in the next subsection.

*3) Fitness of Programs with Small Default Schemas*

We provide fitness statistics for small default schema sizes by presenting such statistics on layers of a Pascal's Pyramid. We are using Pascal's Pyramids layers because they show schemas with the smallest differences in composition next to each other thereby displaying the distances and structural differences between high fitness schemas.

Fig 7 is an example of a single layer of a Pascal's Pyramid suitable for displaying statistics for default schema of size 3.

† Although we will refer to Experiment 2 as a "random walk", it is technically a "random walk with momentum" because executing a "move" or "left" or "right" instruction chosen at random and with equal probability leads to a higher probability of movement in the initial direction and not equal probability of movement in all directions.

‡ The average of 12.3 looks similar to the average cell visits of 123.3. This is not a typo.

Within the top cell will be the statistic of a size 3 schema composed of three “move” instructions. The cell at the center will show statistics for all size 3 schema having one “move”, one “left” and one “right” instruction, such as lmr, rlm or mlr. Note that every cell is different from its neighboring cell’s composition by only one instruction change.

Other Pascal pyramids layers will be similarly arranged with the top, bottom left and bottom right cells showing statistics for schemata with all “move”, all “left” and all “right” instructions respectively.

Pascal Pyramid layers allow us to show changes in schema neighborhood with changes in three instructions (i.e. left, right and move). Note that Pascal Pyramid layers are not Pascal triangles; Pascal triangles, being binomial, would limit visualization of changes to two instructions.

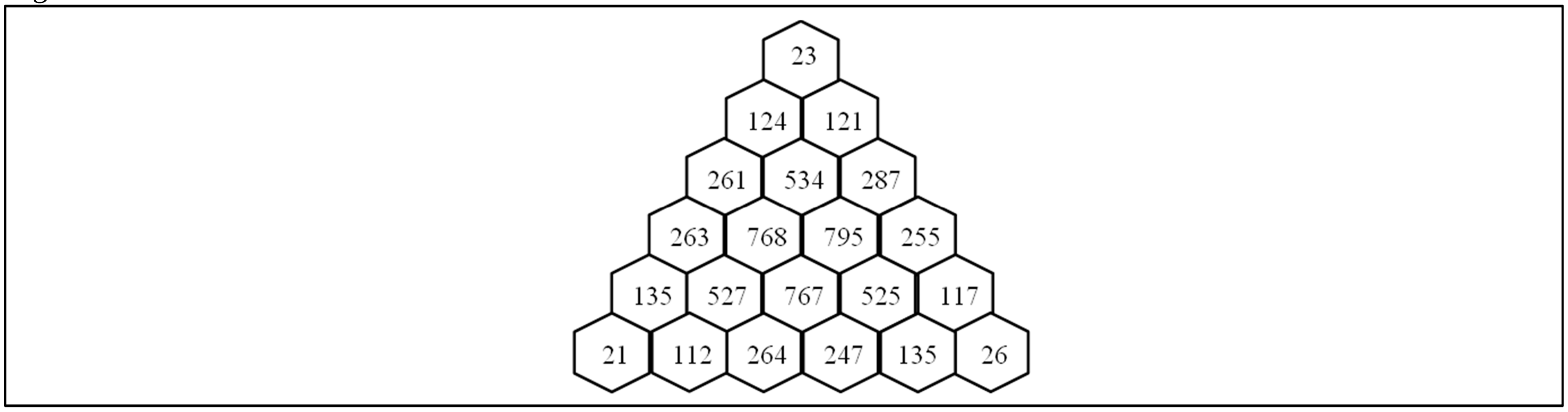

**Fig 8: Population of cells of layer for schema size of 5**

Fig 8 shows the population of various cells for the size 5 schema layer based on the random programs of Experiment 1. The expected size of the population of a schema of size $n$ that corresponds to a cell with composition of $n_m$, $n_l$ and $n_r$ “move”, “left” and “right” instructions is proportional to

$\binom{n}{n_m, n_l, n_r}$, [58]

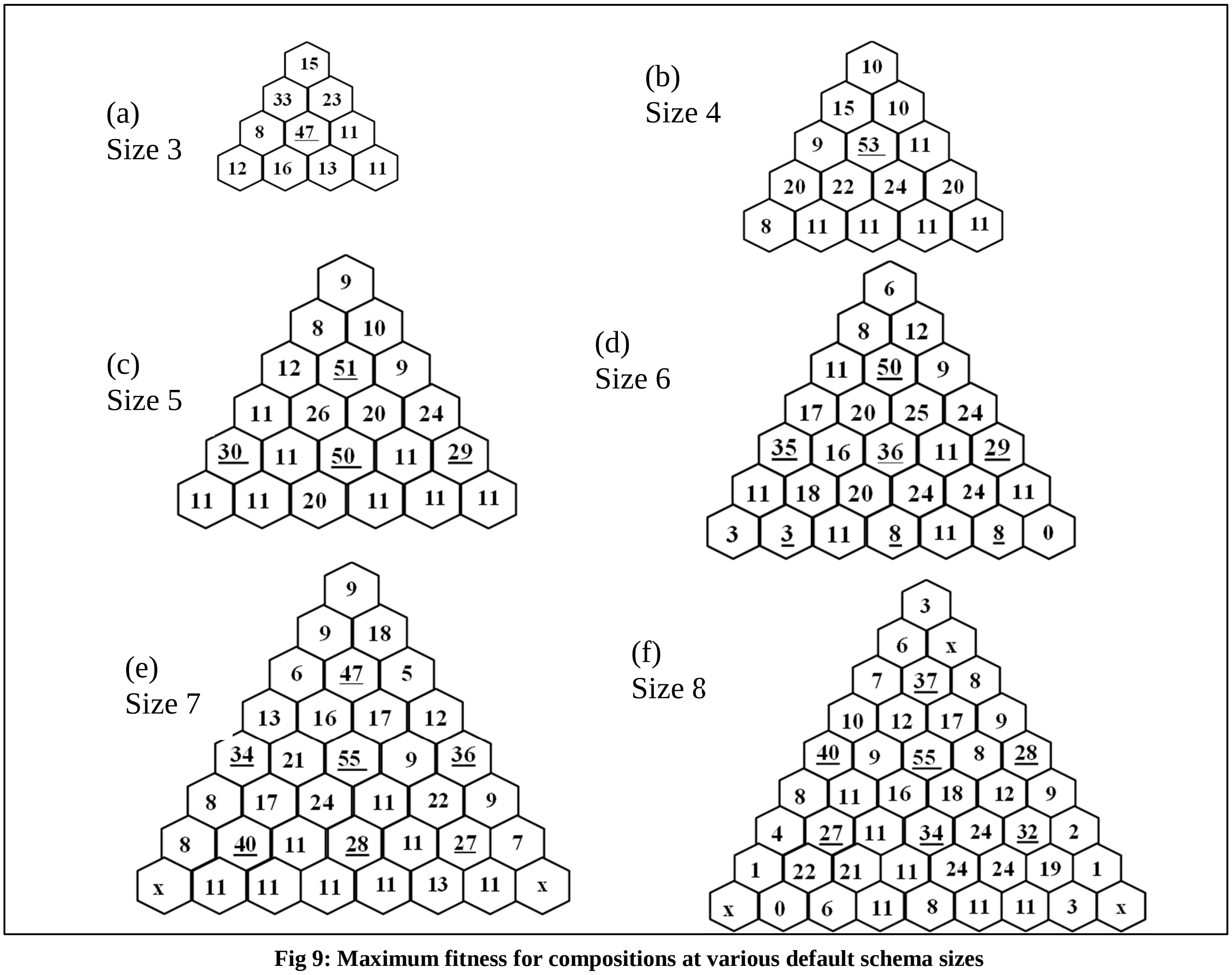

**Fig 9: Maximum fitness for compositions at various default schema sizes**

Fig 9 shows layers for different schema sizes. The layers should be considered as stacked on each other to form a pyramid. Moving from one layer to an adjacent layer would involve adding or removing a “move” or “left” or “right” instruction, thereby increasing or decreasing the size of the schema by one.

Fig 9 shows the maximum fitness scores obtained from the programs of Experiment 1 that have default schema sizes of 3 up to 8. Fig 9(a), for example shows that the maximum fitness we obtained for a size 3 schema having two “move” and a “left” instruction is 33. The “x” in some cells of Fig 9(e) and Fig 9(f) means there were no programs (out of the 90,000 programs of Experiment 1) with those particular default schema composition. Fig 9 shows that fitness peaks are arranged in a pattern; the underlined cells are in most cases fitter than their immediate neighbors. This pattern is explained later in this section.

Fig 10 shows the count of “fit programs” for various default schema compositions. As in [2] we used a minimum fitness score of 25 as our definition of a fit program. 1.3% of the population of Experiment 1 scored 25 or higher. Fig 10 corroborates the pattern from Fig 9 that fit programs are mostly arranged at certain schema compositions which are underlined in the figure.

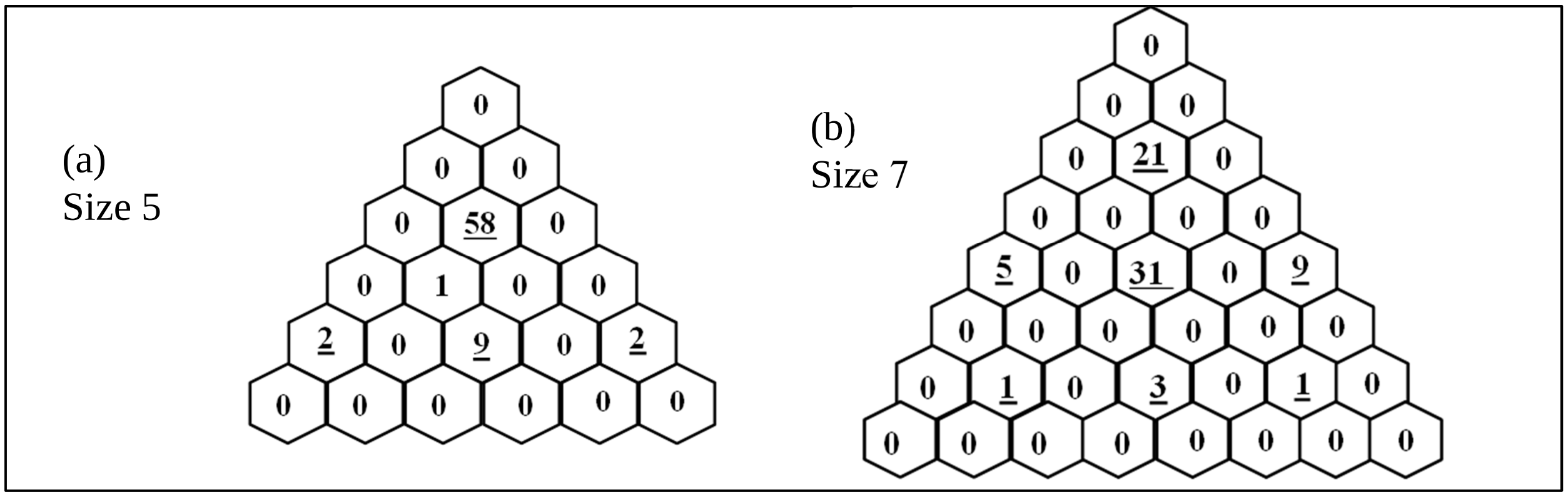

**Fig 10: Number of programs that have fitness of 25 and above at different sizes.**

To explain the pattern of fitness optima for the underlined schema compositions in Fig 9 and Fig 10 we will develop and use the concept of the direction of a schema. We define a schema’s direction as the final direction of an ant on executing that schema’s instructions, relative to the ant’s direction before executing the schema’s instructions.

The four directions a schema can have are:

- Same direction.
- Left.
- Backward.
- Right.

As an example, a schema composed of two “left” and one “move” instructions (irrespective of the order in which those instructions are arranged) will always leave an ant pointing backwards relative to its start direction.

A schema’s direction can be automatically assessed using Eq (IV.1) below:

$$direction \equiv (x - y)\ (mod\ 4) \qquad \text{Eq (IV.1)}$$

Here *x* and *y* represent the number of “left” and “right” instructions in the schema and *mod* is the modulo operation.

If the direction of Eq (IV.1) is zero for a schema, the ant ends up pointing in the original direction it was pointing when it started the instruction sequence. If the direction is 1, 2 or 3, the ant ends up pointing in a direction to the left, backwards, or to the right respectively, relative to its starting direction. As an example the schema mllmr has *x* and *y* as 2 and 1 respectively, leading to a direction of 1, indicating a schema direction of “left”. Schema directions are independent of the order in which the instructions are arranged.

The underlined cells in Fig 9 and Fig 10 (which represent schemas that are fitness optima) are those that leave ant’s pointing in the same direction relative to their start direction. To understand why this is the case, consider that more than 90% of the Santa Fe grid is empty. If an ant does not find food immediately ahead, it uses its default schema sequence. Repeated use of the default schema sequence leads to the ant circling over the same set of empty grid cells, if the default schema’s direction is either left, backward or right. The default schema that leads to visiting many new cells is the one that points in the same direction. As an example a default schema of lmlmm (with a backward direction) will start circle over the same cells after visiting only 6 cells. In contrast a schema of rmlmm (with a same direction) will visit 96 cells before circling over the same set of cells.

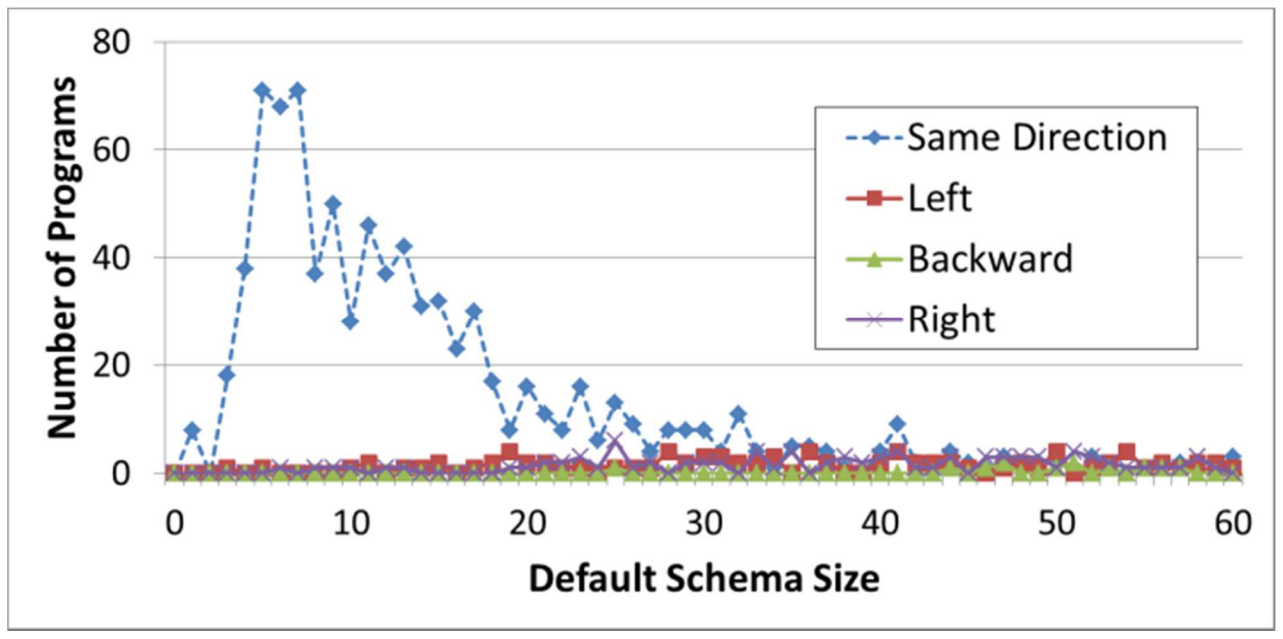

**Fig 11: Number of Fit Programs by Default schema Size and Direction**

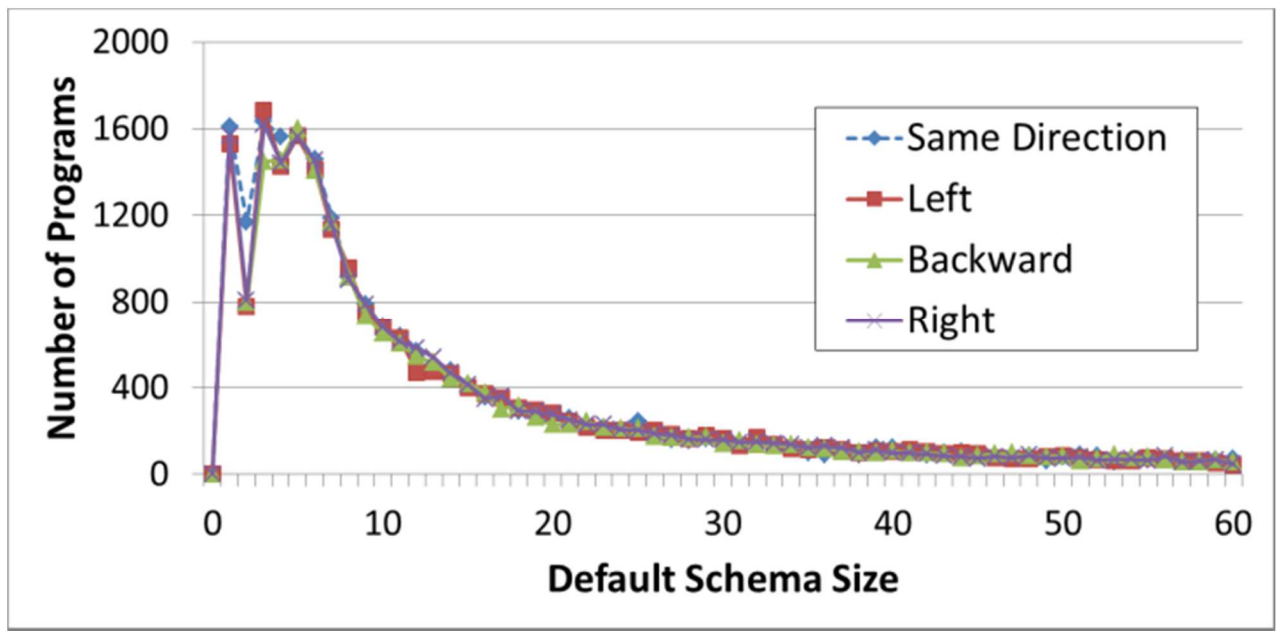

**Fig 12: All Programs by Default schema Size and Direction**

Fig 11 shows a graph of the number of fitter programs plotted by default schema size for all directions. For comparison, Fig 12 shows a similar graph for all programs (not just fitter programs). It can be seen from Fig 11 that for short default schema (of size less than 30), programs that leave the ant pointing in the same direction are disproportionately more fit than those that leave the ant pointing in any other direction. Fig 12 shows there is not such a disproportionate bias for short default schemas in a particular direction when fitness is not considered.

*4) Relating Non-default Schema with Fitness*

We have discussed the relationship between the structure of large and small default schema with fitness in the previous two subsections. That relationship is strong because the default schema is used for the empty cells of the grid and more than 90% of the grid is empty.

There are also schemas that respond to other trail conditions other than not finding food. Fig 13 shows the variation of the average fitness of programs with the number of different schemas in the program. Fig 13 is from the random population of Experiment 1. Fig 13 shows that the average fitness of randomly generated programs increases with the number of schemas (or different responses to trail conditions) they have.

This finding is not surprising as high fitness is achieved when programs visit many cells. More diverse responses to different trail conditions make it less likely that a program will to be stuck cycling over a limited number of cells.

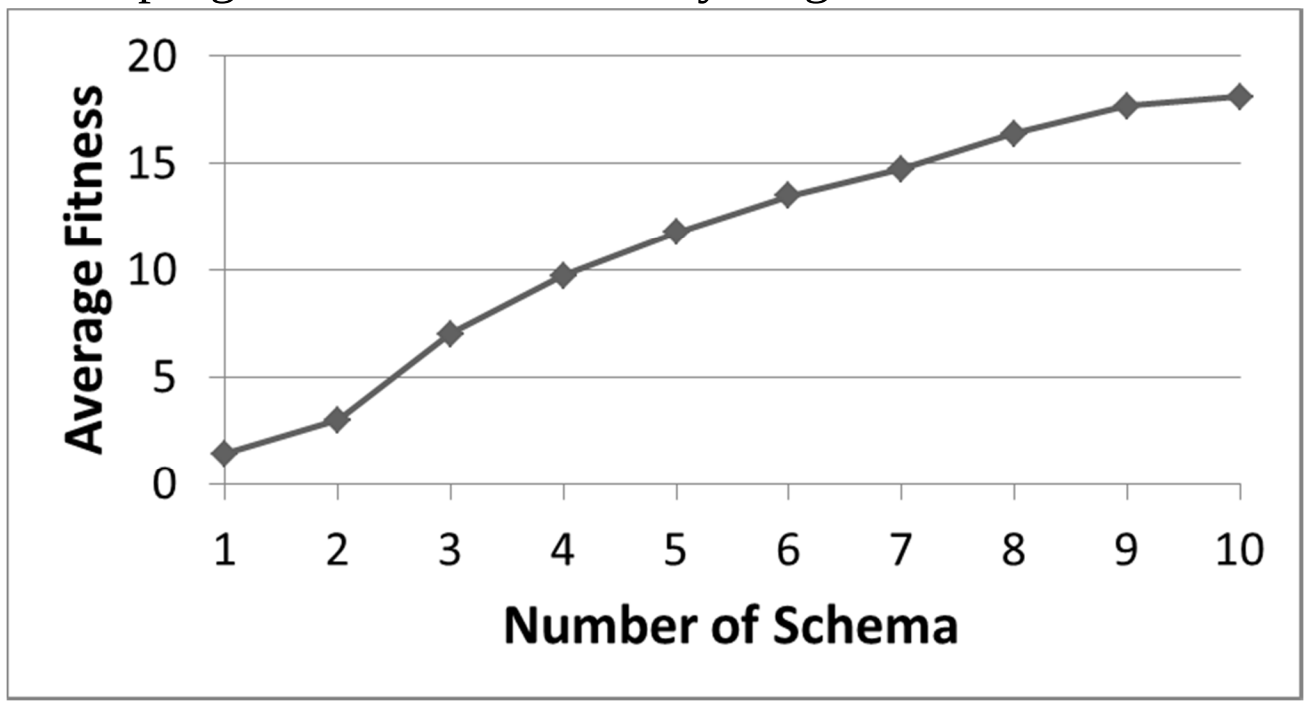

**Fig 13: Average fitness of randomly generated programs with different numbers of schema**

*5) Discussion of Findings in Relating Fitness and Structure*

It is important to note that the findings in this section apply only to randomly generated programs that have not been subjected to any evolutionary process. As a result the findings are not limited to GP, but general to the representation we are using. Findings that are relevant to the GP evolutionary process are discussed in a subsequent section.

We have determined that average fitness is related to program structure. This relation is based on program default schemata determining program behavior, especially the number of different cells a program visits.

Most high fitness programs are those that have either:

- Short default schema that leave ants pointing in their original direction, or;
- Long default schema irrespective of direction.

This fact is observable from Fig 11. Its reason is that to avoid cycling over a small set of the same cells, ants need to do either:

- Execute the many instructions in a long default schema, thereby approximating a random walk on the grid§.
- Execute the few instructions in a short default schema, and visit new cells by repetition of the default schema instructions. The schema that cause ants to end pointing in the same direction they started are those that visit many new cells.

Observation of Fig 9 and Fig 10 show that the short fit schema that characterize local optima are arranged in a regular pattern. Fig 14 shows the pattern of schema directions for schemas of size 5. It can be seen from Fig 14 that schemas with the same direction have a distance of at least two instruction changes between them. This means that we need at least two instruction changes to go from one optimum to another on the same layer. If we overlay the layers in Fig 9 however, we see that we can move from one optimum to another by adding or removing a single “move” instruction. All other changes from optima to optima require at least two instruction changes.

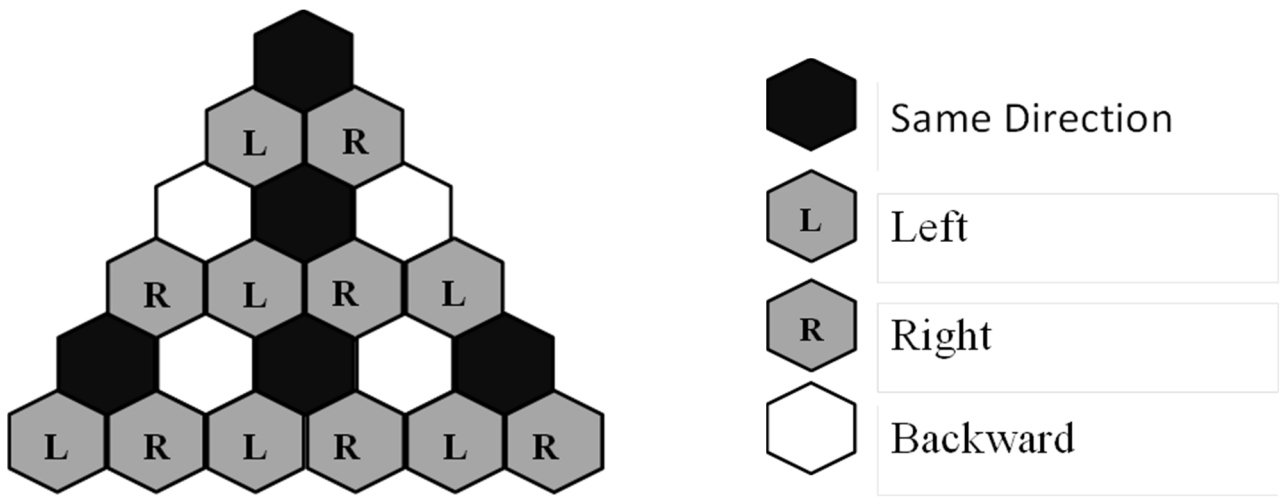

**Fig 14: Schema Directions of schemas of size 5**

Another finding (from Fig 6) is that the average fitness of programs increases with the size of their default schema. This is because most programs with short default schema have low fitness and a few have high fitness. Programs with large schema sizes tend to have a lower spread in their fitness (observable from Fig 6) due to their approximating a random walk.

**Table 1: Parameter settings for Experiment 3**

| **Parameter** | **Value** |
|---|---|
| Population | 500 |
| Generations | 50 |
| Terminal Set | Left, Right, Move |
| Non-Terminal Set | IfFoodAhead, Prog2, Prog3 |
| Success Predicate | Fitness score = 89 |
| Initial Population generation method | Ramped half-and-half with depth of 2 to 6 |
| Max depth | 17 |
| Selection (for both crossover and reproduction) | Tournament (size=7) |
| Crossover | 0.9 |
| Reproduction | 0.1 |
| Starting Energy | 600 |

Langdon and Poli [2] found that the average fitness of randomly generated programs increases with their size while the variance of fitness is reduced with increase in size. We consider this a result of the fact that large programs tend to have large default schemas and small programs tend to have short default schemas.

## V. The Schema Structure of Solutions

In the previous section we discussed structures that make ants fit. It is important that we also analyze the structures of solutions (i.e. ants that consume all food on the trail).

### *A. Schema of Evolved and Random Solutions*

To observe the structure of solutions obtained through evolution we ran 6000 replicates of GP on the Santa Fe ant problem for

§ Large randomly generated or evolved programs, that follow the trail without using the implicit iteration mechanism are possible in principle; However they are rare as we did not find any example on extensive examination of solutions. Solutions are discussed in the next section.

50 generations, using the settings from Table 1. The settings in Table 1 include default settings from Koza [8]. We will refer to this experiment as Experiment 3. We obtained 617 solutions ranging in tree size from 13 nodes to 509 nodes. We now analyze the default schema of these solutions.

Fig 15 shows the distribution of the number of solutions of default schema size 5 and size 7 arranged on Pascal pyramids. Solutions of default schema sizes 5 and 7 made up 95.3% of all solutions. The remaining solutions range in default schema size from 9 to 32. As such the majority of evolved solutions have short default schema sizes. All solutions had default schema with "same direction", and it can be seen from Fig 15 that the solutions occupy a subset of the schema compositions for "same direction" schema.

While Fig 15 gives a default schema analysis of solutions that were evolved, we also conducted default schema analysis of randomly generated solutions. We analyzed the default schema of the 3916 randomly generated solutions from Langdon and Poli [2]. 98.2% of the solutions had a schema of size 5 and the remaining solutions had a schema size of 7. Fig 17 shows the default schema composition of all the random solutions. Here again the solutions occupy a subset of the schema compositions for "same direction".

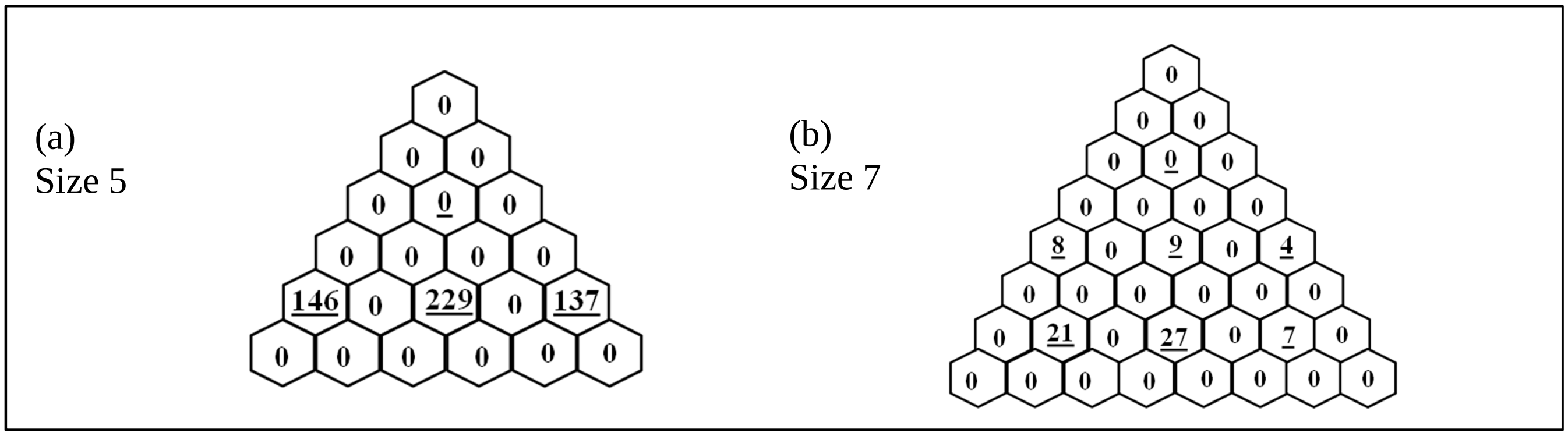

**Fig 15: Number of Evolved Solutions with Default Schema of size 5 and 7**

*B. Food Trail Structure and Default Solution Schema*

There are four mutually exclusive possibilities for how the food trail displayed in Fig. 1 continues forward from any point on it. These are:

- There is food on the spot ahead;
- There is food (a) immediately to the left, or (b) immediately to the right, of the current position;
- There is a trail gap i.e. the trail continues forward with no food in the immediate surrounding cells.

Assume an ant is facing the trail and encounters a trail gap. In order for the ant to correctly recognize and follow the trail gap, it has to ensure that there is no food to its immediate left or right, and then move in the direction of the trail. This means that for correctly following a trail gap with the minimum number of instructions, the ant has to turn to the left and then to the right of the trail (or vice versa), ensure there are no food in either direction and then point in the original direction of the trail and move in that direction.

The minimum number of instructions required for correctly following the trail gap is five. There are four sequences of five instructions that turn to both the left and right of the trail direction and then move in the original trail direction; these are listed in Table 2. Each of these sequences gives another valid sequence (that recognizes trail gaps) on *rotation*. A rotation is one or more transformations of a sequence of instructions:

$x_0 x_1 \cdots x_{n-1} x_n \rightarrow x_n x_0 \cdots x_{n-2} x_{n-1}$ .

For example rllrm rotated twice results in the sequence rmrll.

The sequences in Table 2 (and their rotations) are important because all size 5 solutions (both evolved and randomly generated) use one of them as their default schema sequence. All size 5 solutions in Fig 15(a) and Fig 17(a) use one of these sequences, or some rotation of these sequences.

TABLE 2: MINIMUM DEFAULT SCHEMA SEQUENCE TO ASCERTAIN A TRAIL GAP

| Sequence: | Instructions | | | | |
|---|---|---|---|---|---|
| 1 | left | left | left | left | move |
| 2 | left | right | right | left | move |
| 3 | right | right | right | right | move |
| 4 | right | left | left | right | move |

In the majority of cases (72% for evolved solutions, 99% for randomly generated solutions), size 7 sequence happen to be one of the size 5 sequences indicated in Table 2, with a left and right (or right and left) inserted somewhere within e.g. $lll\boldsymbol{rl}lm$.

### *C. Synchronization Requirements of Solutions*

For an ant to successfully navigate the trail it needs some mechanism to keep track of the trail's direction. There are no program primitives that directly implement this memory function. We now explain that this memory function is established in most solutions as a synchronization requirement between all schema of a solution. In most cases this synchronization is achieved by all schema having compositions that orient the ant in a common direction relative to the trail. The four possible schema orientations are forward, backward, left and right[**].

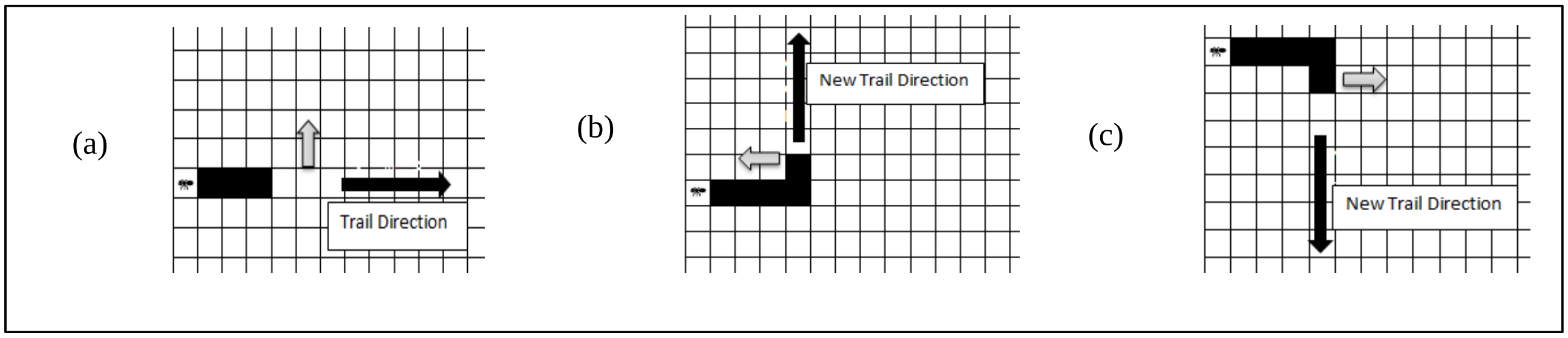

**Fig 16: Trails for measuring the orientation of solution schema**

To measure the orientations of the schema of solutions we ran each solution on the three trails shown in Fig 16. In Fig 16 (a) the trail goes forward eastwards and then ends. We find the orientation for default schema on this trail by noting the direction of the ant (relative to the trail's direction) after consuming all three food items and going over an empty cell. In Fig 16 (b) and (c), the trail goes forward and then takes a left turn and right turn respectively. We find the orientation of the response to left and right turns on these respective trails by finding the final relative direction of the ant on consuming all food. As examples the white arrows in Fig 16 shows the direction that an ant that always orients to the left of the trail, (such as the program in Fig 2(a) ) would point.

The orientations of solution schemas were measured for all solutions from experiment 3. These orientations were also measured for the random solutions from Langdon and Poli [2]. 93% of evolved solutions and 99.9% of the random solutions have all their schema oriented in the same direction. It can therefore be concluded that the "memory" mechanism used by most solutions is to orient the ant towards a certain direction at the end of each iteration of its code. This orientation is used for all conditions on the trail so that the ant always knows its direction relative to the trail's direction.

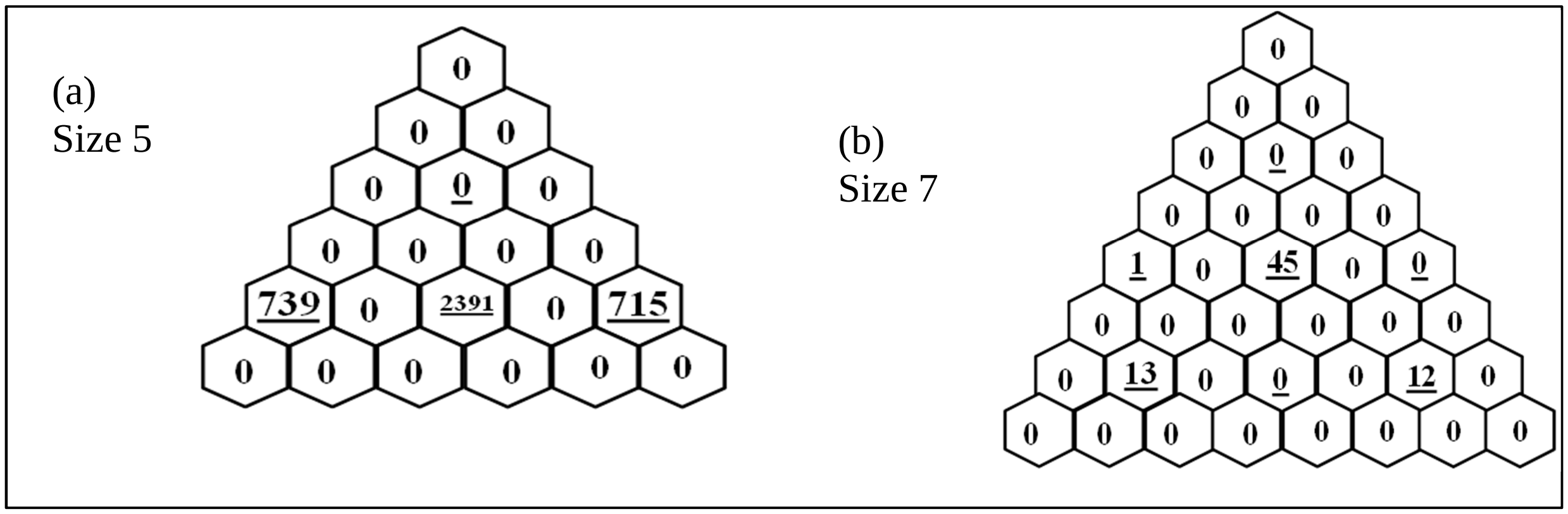

**Fig 17: Count of Randomly Generated Solutions**

## VI. How Variations Search the Landscape

We used the results from Experiment 3 (i.e. 6000 trials of GP on the Santa Fe ant problem for 50 generations, using the settings from Koza [8]), to understand how crossover searches the landscape. The settings include a 90% crossover rate. We recorded the structures and fitness scores of the fittest program at each generation for which there is a fitness increase.

We now analyze how the composition of the default schema changes from the current fittest program's structure to the next fittest program's structure. Fig 18 shows the changes in default schema (that result in an increase in maximum fitness) for various default schema sizes.

[**] Schema orientation should not be confused with schema direction. The schemas direction was defined in Section IV.B as the final direction of an ant on executing that schema's instructions, relative to the ant's direction before executing the schema's instructions.

Next Default Schema (Move-Left-Right)

Current Default Schema (Move-Left-Right)

| | | 1 | 2 | 3 | | | 4 | | | | | 5 | | | | | | | |
|---|---|---|---|---|---|---|---|---|---|---|---|---|---|---|---|---|---|---|---|
| | | 1-0-0 | 2-0-0 | 1-1-1 | 2-0-1 | 2-1-0 | 1-1-2 | 1-2-1 | 1-3-0 | 2-1-1 | 3-1-0 | 1-0-4 | 1-2-2 | 1-4-0 | 2-0-3 | 2-1-2 | 2-2-1 | 2-3-0 | 3-1-1 |
| 1 | 1-0-0 | | 5 | 3 | | | | | | 11 | | 2 | | 1 | | | 1 | | 12 |
| 2 | 2-0-0 | 2 | | 4 | | | | | | 9 | | 2 | 1 | | | | | | 7 |
| 3 | 1-1-1 | 5 | 5 | | | | | | | 76 | | 5 | 13 | 4 | 1 | | | | 103 |
| | 2-0-1 | | | | | | | | | | | | | | | | | | |
| | 2-1-0 | | | | | | | | | | | | | | | | | | |
| 4 | 1-1-2 | | | | | | | | | 1 | | | | | | | | | |
| | 1-2-1 | | | | | | | | | | | | | | | | | | |
| | 1-3-0 | | | | | | | | | | | | | | | | | | |
| | 2-1-1 | 8 | 7 | 63 | | | | | | | | 5 | 16 | 6 | | 1 | 2 | 2 | 307 |
| | 3-0-1 | | | | | | | | | | | | | | | | | | |
| | 3-1-0 | | | | | | | | | | | | | | | | | | 1 |
| 5 | 0-3-2 | | | | | | | | | | | | | | | | | | |
| | 1-0-4 | | 1 | 3 | | | | | | 4 | | | 5 | 3 | | | | | 9 |
| | 1-2-2 | 2 | 2 | 17 | | | | | | 13 | | 6 | | 7 | | | 1 | | 23 |
| | 1-4-0 | | | 1 | | | | | | 3 | | 3 | 3 | | | | | | 4 |
| | 2-0-3 | | | | | | | | | | | 2 | | | | | | | 2 |
| | 2-1-2 | | | | | | | | | | | | | | | | | | 1 |
| | 2-2-1 | | | | | | | | | 1 | | | | | | | | | 1 |
| | 2-3-0 | | | | | | | | | | | | | 2 | | | | | |
| | 3-1-1 | 12 | 8 | 91 | | 1 | | | | 279 | | 8 | 31 | 6 | 2 | 2 | 3 | 3 | |
| 6 (current schema size) | 1-1-4 | | | | | | | | | | | | | | | | | | 1 |
| | 1-2-3 | | | | | | | | | | | | | | | | | | 1 |
| | 1-3-2 | | | | | | | | | | | | | | | | | | 1 |
| | 1-4-1 | | | | | | | | | | | | | | | | | | 1 |
| | 2-0-4 | | | 1 | | | | | | 1 | | 5 | 3 | | | | | | 8 |
| | 2-2-2 | 4 | | 16 | | | | | | 62 | | 3 | 27 | 1 | | | | | 67 |
| | 2-3-1 | 1 | | | | | | | | | | | | | | | | | 1 |
| | 2-4-0 | 1 | 1 | 3 | | | 1 | | | 5 | | 1 | | 7 | | | | | 11 |
| | 3-0-3 | | | | | | | | | | | | | | | | | | |
| | 3-1-2 | | | 1 | | | | | | | | | 1 | | | | | | |
| | 3-2-1 | | | | | | | | | 1 | | | | | | | | | 2 |
| | 4-1-1 | 4 | 2 | 24 | | | | | 1 | 117 | | 12 | 11 | 5 | | 2 | | | 211 |
| 7 | 1-1-5 | | | 2 | | | | | | 1 | | 55 | 2 | | | | | | 5 |
| | 1-3-3 | | 1 | 5 | | | | | | 4 | | 4 | 72 | 11 | | 2 | | | 12 |
| | 1-5-1 | | | 3 | | | | | | 3 | | | 11 | 46 | | | | | 6 |
| | 2-0-5 | | | | | | | | | 1 | | | | | | | | | |
| | 2-1-4 | | | | | | | | | | | | | | | | | | 2 |
| | 2-2-3 | | | | | | | | | 3 | | | 1 | | | 1 | | | 2 |
| | 2-3-2 | | | 1 | | | | | | 1 | | | | | | | 1 | | 1 |
| | 2-4-1 | | | | | | | | | 1 | | 1 | | | | | | | 1 |
| | 2-5-0 | | | | | | | | | | | | | | | | 1 | | |
| | 3-0-4 | | | | | | | | | 6 | | 14 | 1 | | | | | | 13 |
| | 3-1-3 | | | | | | | | | | | | | | | | | | |
| | 3-2-2 | 2 | 3 | 24 | | | | | | 71 | | 7 | 39 | 3 | 1 | 1 | 1 | | 190 |
| | 3-3-1 | | | | | | | | | | | | | | | | | | |
| | 3-4-0 | 1 | | 4 | | | | | | 11 | | | | 19 | | | | | 9 |
| | 4-1-2 | | | | | | | | | 1 | | | | | | | | | 2 |
| | 4-2-1 | | | | | | | | | 1 | | | | | | | | | |
| | 4-3-0 | | | | | | | | | | | | | | | | | | 1 |
| | 5-1-1 | 2 | 1 | 29 | | | | | | 67 | | 4 | 10 | 2 | 1 | 1 | | | 210 |
| | 6-0-1 | | | | | | | | | | | | | | | | | | 1 |
| Total Next Default Schema | | 106 | 78 | 660 | 4 | 1 | 1 | 1 | 1 | 1498 | 2 | 228 | 414 | 201 | 11 | 12 | 15 | 12 | 2546 |

| | | 6 | | | | | | | | | | |
|---|---|---|---|---|---|---|---|---|---|---|---|---|
| | | 1-1-4 | 1-2-3 | 1-3-2 | 1-4-1 | 2-0-4 | 2-2-2 | 2-4-0 | 3-0-3 | 3-1-2 | 3-2-1 | 4-1-1 |
| 1 | 1-0-0 | | | | | 2 | 2 | 2 | | | | 6 |
| 2 | 2-0-0 | | | | | | 1 | | | | | 1 |
| 3 | 1-1-1 | | | | | | 8 | 3 | | | | 25 |
| | 2-0-1 | | | | | | 1 | | | | | |
| | 2-1-0 | | | | | | | | | | | |
| 4 | 1-1-2 | | | | | | | | | | | |
| | 1-2-1 | | | | | | | | | | | |
| | 1-3-0 | | | | | | | | | | | 1 |
| | 2-1-1 | | | | | 3 | 31 | 2 | | | 1 | 133 |
| | 3-0-1 | | | | | | 1 | | | | | |
| | 3-1-0 | | | | | | | | | | | 1 |
| 5 | 0-3-2 | | | | | | | | | | | |
| | 1-0-4 | | | | | 2 | 1 | | | | | 6 |
| | 1-2-2 | | | | | | 13 | | | | 1 | 5 |
| | 1-4-0 | | | | | | 1 | 3 | | | | 4 |
| | 2-0-3 | | | | | | | | | | | |
| | 2-1-2 | | | | | | | | | | | |
| | 2-2-1 | | | | | | | | | | 1 | |
| | 2-3-0 | | | | | | | | | | | 1 |
| | 3-1-1 | | | 1 | | 4 | 43 | 8 | | 1 | 2 | 243 |
| 6 (current schema size) | 1-1-4 | | | | | | | | | | | |
| | 1-2-3 | | | | | | | | | | | |
| | 1-3-2 | | | | | | | | | | | |
| | 1-4-1 | | | | | | | | | | | |
| | 2-0-4 | | | | | | 2 | | | | | 4 |
| | 2-2-2 | | | | | 2 | | 4 | | 1 | | 20 |
| | 2-3-1 | | | | | | | | | | | |
| | 2-4-0 | | | | | | 3 | | | | | 2 |
| | 3-0-3 | | | | | | | 1 | | | | |
| | 3-1-2 | | | | | | | | | | | |
| | 3-2-1 | | | | | | | | | | | 1 |
| | 4-1-1 | | | | | 2 | 16 | 2 | | 1 | | |
| 7 | 1-1-5 | | | | | | | | | | | 3 |
| | 1-3-3 | | | | | | 4 | | | | | 5 |
| | 1-5-1 | | | | | 1 | 1 | | | | | 4 |
| | 2-0-5 | | | | | | | | | | | |
| | 2-1-4 | | | | | | | | | | | |
| | 2-2-3 | | | | | | | | | | | |
| | 2-3-2 | | | | | 1 | | | | | | |
| | 2-4-1 | | | | | | | 1 | | | | |
| | 2-5-0 | | | | | | | | | | | |
| | 3-0-4 | | | | | 3 | | | 1 | | | 8 |
| | 3-1-3 | | | | | | | | | | | |
| | 3-2-2 | | 1 | | | | 38 | 2 | | | 1 | 43 |
| | 3-3-1 | | | | | | | | | | | |
| | 3-4-0 | | | | | | 3 | 5 | | | | 8 |
| | 4-1-2 | | | | | | | | | | | |
| | 4-2-1 | | | | | | | | | | | |
| | 4-3-0 | | | | | | | | | | | |
| | 5-1-1 | | | | | 2 | 11 | 3 | | | 2 | 131 |
| | 6-0-1 | | | | | | | | | | | |
| Total Next Default Schema | | 2 | 2 | 1 | 1 | 54 | 440 | 84 | 1 | 6 | 15 | 1373 |

| | | 7 (next schema size) | | | | | | | | | | | | | | | |
|---|---|---|---|---|---|---|---|---|---|---|---|---|---|---|---|---|---|
| | | 1-1-5 | 1-3-3 | 1-5-1 | 2-1-4 | 2-2-3 | 2-3-2 | 2-4-1 | 3-0-4 | 3-1-3 | 3-2-2 | 3-3-1 | 3-4-0 | 4-1-2 | 4-2-1 | 4-3-0 | 5-1-1 |
| 1 | 1-0-0 | | 1 | | | | | 1 | | | 7 | | 1 | | | | 6 |
| 2 | 2-0-0 | | | | | | | | | | 2 | | 1 | | | | 4 |
| 3 | 1-1-1 | 4 | 7 | 5 | | | 1 | | 1 | | 27 | | 4 | | | | 38 |
| | 2-0-1 | | | | | | | | | | | | | | | | |
| | 2-1-0 | | | | | | | | | | | | | | | | |
| 4 | 1-1-2 | | | | | | | | | | | | | | | | |
| | 1-2-1 | | | | | | | | | | | | | | | | |
| | 1-3-0 | | | | | | | | | | | | | | | | |
| | 2-1-1 | 7 | 7 | 6 | 1 | 1 | | | 7 | | 63 | | 13 | | 1 | | 98 |
| | 3-0-1 | | 1 | | | | | | | | | | | | | | |
| | 3-1-0 | | | | | | | | | | | | | | | | |
| 5 | 0-3-2 | | | | | | | | | | | | | | | | |
| | 1-0-4 | 10 | 3 | 2 | | | | | 5 | | 1 | | | | | | 4 |
| | 1-2-2 | 2 | 10 | 4 | | 1 | | 1 | 2 | | 17 | | 1 | | | | 7 |
| | 1-4-0 | | 3 | 9 | | | | | | | 1 | | 5 | | | | 1 |
| | 2-0-3 | 1 | | | | | | | 1 | | | | | | | | |
| | 2-1-2 | | 2 | | | 4 | | | | | | | | | | | |
| | 2-2-1 | | 2 | | | 1 | | | | | | | | | | | 1 |
| | 2-3-0 | | | | | | | | | | 1 | | | | | | 1 |
| | 3-1-1 | 6 | 11 | 13 | 2 | 1 | | 1 | 11 | | 127 | | 16 | 4 | | 2 | 263 |
| 6 (current schema size) | 1-1-4 | | | | | | | | | | | | | | | | |
| | 1-2-3 | | | | | | | | | | | | | | | | |
| | 1-3-2 | | | | | | | | | | | | | | | | |
| | 1-4-1 | | | | | | | | | | | | | | | | 1 |
| | 2-0-4 | 1 | | 1 | | | 1 | | 1 | | | | | | | | 2 |
| | 2-2-2 | 2 | 11 | | | 1 | | 1 | 2 | | 43 | | | | | | 16 |
| | 2-3-1 | | | | | | | | | | | | | | | | |
| | 2-4-0 | | | 2 | | | | | | | 2 | | 9 | | | | 1 |
| | 3-0-3 | | | | | | | | | | | | | | | | |
| | 3-1-2 | | | | | | | | | | | | | | | | |
| | 3-2-1 | | | | | | | | | | | | | | | | 1 |
| | 4-1-1 | 6 | 2 | 6 | | | | 1 | 8 | | 46 | | 4 | | | | 145 |
| 7 | 1-1-5 | | 6 | | | | | | | | 2 | | | | | | 2 |
| | 1-3-3 | 7 | | 9 | | | | | | | 4 | | | | | | 7 |
| | 1-5-1 | 1 | 9 | | | | | | | | 1 | | 1 | | | | 1 |
| | 2-0-5 | | | | | | | | | | | | | | | | |
| | 2-1-4 | | | | | | | | 1 | | | | | | | | 1 |
| | 2-2-3 | | 1 | | | | | | | 1 | 1 | | | | | | |
| | 2-3-2 | | | | | | | | | | | | | | | | 1 |
| | 2-4-1 | | | | | | | | | | | | | | | | |
| | 2-5-0 | | | | | | | | | | | | | | | | |
| | 3-0-4 | 1 | 1 | | | 1 | | | | | 4 | | 1 | | | | 4 |
| | 3-1-3 | | | | | | | | | | | | | | | | |
| | 3-2-2 | 5 | 9 | 1 | | | | | 4 | | | | 7 | | | | 43 |
| | 3-3-1 | | | | | | | | | | | | 1 | | | | |
| | 3-4-0 | 1 | | 1 | | | | | 2 | | 6 | | | | | | 5 |
| | 4-1-2 | | 1 | | | | | | | | | | | | | | 1 |
| | 4-2-1 | | | | | | | | | | | | | | | | 1 |
| | 4-3-0 | | | | | | | | | | 1 | | | | | | |
| | 5-1-1 | 1 | 8 | | | | 1 | | 4 | | 30 | | 3 | 1 | | | |
| | 6-0-1 | | | | | | | | | | | | | | | | |
| Total Next Default Schema | | 124 | 230 | 155 | 4 | 20 | 8 | 9 | 145 | 1 | 986 | 1 | 169 | 9 | 1 | 2 | 1475 |

| | | 8 | | | | | | | | | | | | | Total Current Default Schema |
|---|---|---|---|---|---|---|---|---|---|---|---|---|---|---|---|
| | | 1-3-4 | 1-4-3 | 1-5-2 | 1-6-1 | 2-1-5 | 2-3-3 | 2-5-1 | 3-1-4 | 3-2-3 | 3-3-2 | 3-4-1 | 4-0-4 | 4-2-2 | |
| 1 | 1-0-0 | | | | | | | | | 1 | | | | 3 | 138 |
| 2 | 2-0-0 | | | | | | | | | | | | 1 | 3 | 69 |
| 3 | 1-1-1 | | | | 1 | 2 | 5 | 1 | | | | | | 13 | 663 |
| | 2-0-1 | | | | | | | | | | | | | | 6 |
| | 2-1-0 | | | | | | | | | | | | | | 1 |
| 4 | 1-1-2 | | | | | | | | | | | | | | 1 |
| | 1-2-1 | | | | | | | | | | | | | | 1 |
| | 1-3-0 | | | | | | | | | | | | | | 1 |
| | 2-1-1 | | | | | 1 | 10 | 1 | 1 | | | | 4 | 51 | 1527 |
| | 3-0-1 | | | | | | | | | | | | | | 2 |
| | 3-1-0 | | | | | | | | | | | | | | 3 |
| 5 | 0-3-2 | | | | | | | | | | | | | | 1 |
| | 1-0-4 | 1 | | | | | | | | | | | 1 | | 101 |
| | 1-2-2 | | | | | 1 | 2 | 1 | | | | | 1 | 4 | 235 |
| | 1-4-0 | | | | | | | | | | | | 1 | 1 | 67 |
| | 2-0-3 | | | | | | | | | | | | | | 7 |
| | 2-1-2 | | | | | | | | | | | | | | 13 |
| | 2-2-1 | | | | | | | | | | | | | | 12 |
| | 2-3-0 | | | | | | | | | | | | | | 11 |
| | 3-1-1 | | | | | | 7 | 4 | 1 | 1 | 2 | 2 | 5 | 78 | 2406 |
| 6 (current schema size) | 1-1-4 | | | | | | | | | | | | | | 2 |
| | 1-2-3 | | | | | 1 | | | | | | | | | 4 |
| | 1-3-2 | | | | | | | | | | | | | | 2 |
| | 1-4-1 | | | | | | | | | | | | | | 3 |
| | 2-0-4 | | | | | | | | | | 1 | | 2 | | 60 |
| | 2-2-2 | | | | | 1 | 7 | 1 | 1 | | 2 | | | 20 | 542 |
| | 2-3-1 | | | | | | | | | | | | | | 2 |
| | 2-4-0 | | | | 1 | | 1 | 2 | | | | | | | 101 |
| | 3-0-3 | | | | | | | | | | | | | 1 | 2 |
| | 3-1-2 | | | | | | | | | | | | | | 5 |
| | 3-2-1 | | | | | | | | | | | | | | 10 |
| | 4-1-1 | | | | | 2 | 6 | 1 | | | | | 4 | 41 | 1294 |
| 7 | 1-1-5 | | | | | | | | | | | | | 2 | 112 |
| | 1-3-3 | | | | | | 1 | | | | | | | 5 | 207 |
| | 1-5-1 | | | | | | | 1 | | | | | | 2 | 129 |
| | 2-0-5 | | | | | | | | | | | | | | 1 |
| | 2-1-4 | | | | | | | | | | | | | | 7 |
| | 2-2-3 | | | | | | | | | 2 | | | | | 18 |
| | 2-3-2 | | | | | | | | | | | | 1 | | 9 |
| | 2-4-1 | | | | | | | | | | | 1 | | | 11 |
| | 2-5-0 | | | | | | | | | | | | | | 1 |
| | 3-0-4 | | | | | 1 | | | | | | | 8 | 2 | 140 |
| | 3-1-3 | | | | | | | | | | | | | 1 | 1 |
| | 3-2-2 | | | | | 3 | 3 | 3 | | | 1 | 1 | 3 | 45 | 1014 |
| | 3-3-1 | | | | | | | | | | | | | | 1 |
| | 3-4-0 | | | | | | | 1 | | | | | | 2 | 159 |
| | 4-1-2 | | | | | | | | | | | | | | 9 |
| | 4-2-1 | | | | | | | | | | | | | | 2 |
| | 4-3-0 | | | | | | | | | | | | | | 2 |
| | 5-1-1 | | | | | | | 1 | | | 2 | | | 21 | 1237 |
| | 6-0-1 | | | | | | | | | | | | | | 1 |
| Total Next Default Schema | | 1 | 2 | 1 | 2 | 31 | 101 | 42 | 4 | 5 | 14 | 5 | 70 | 672 | |

**Fig 18: Change in default schema of fittest programs**

The default schema composition is shown in the format M-L-R, where M, L and R are the number of “move”, “left” and “right” instructions in the default schema. Fig 18 shows, for example, that there were 46 occasions when a population whose fittest program had a size 7 default schema with M-L-R composition of 1-5-1 was replaced by a future population having a program of higher fitness and default schema size of 5 and M-L-R of 1-4-0.

Cases when the default schema remains the same with increase in fitness are not shown in Fig 18. The grid of Fig 18 is limited to what can reasonably be shown on a printed page. The totals and their charts (along the right and bottom margins) are however inclusive of all changes that occurred, not just those detailed for the schemas shown on Fig 18.

The totals, and their charts, along the margins of Fig 18 show peaks for such M-L-R compositions as 1-1-1, 2-1-1, 3-1-1 and 3-2-2. This means that many of the changes of default schema compositions that occur do so between compositions that meet the characteristic we identified for high fitness programs in Section IV; i.e. most fittest current and future programs have a M-L-R composition of a “same direction” schema. Eq (IV.1) evaluate to 0 for the default schema of these programs.

It can be deduced from Fig 18 that the primary variations that lead to higher fitness are those that do not alter Eq (IV.1) while changing composition. The variations include the following and their combinations:

- The addition or deletion of one or more “move” commands. An example of such an addition is a change from M-L-R of 1-1-1 to 3-1-1. An example of a deletion is change from 3-1-1 to 2-1-1.
- The addition or deletion of an equal number of right and left commands. An example of such an addition is change from 2-1-1 to 2-2-2. An example of a deletion is change from 3-2-2 to 3-1-1.

Other than the addition or deletion of a single move command, all other variations that lead to high fitness require a change to more than one instruction. The most common variation in Fig 18 involves the addition or deletion of single move instructions such as from composition 2-1-1 to 3-1-1 and vice versa. Empty cells in Fig 18 represent situations where there was not a change that resulted in a program of higher fitness.

Fig 18 shows that the most prominent source and destination schema is of size 5 with three move, one left, and one right commands. This schema is key to understanding what happens to evolutionary runs that do not lead to solutions. Most runs that do not find solutions end up with their fittest program having the default schema composition of 3-1-1.

We showed the default schema composition of solutions on Pascal pyramids in Fig 15 and Fig 17. These solutions have schema compositions 1-2-2, 1-0-4 and 1-4-0 for schemas of size 5. It is instructive to note from Fig 18 the schemas that change into the above mentioned solution schemas most involve the change of two instructions, with, for instance, 1-1-5 and 1-5-1 going to 1-0-4 and 1-4-0 respectively.

### *A. How Santa Fe Ants Evolve*

Based on our findings, the way Santa Fe ant programs evolve can be described as follows:

A common property of fit programs that are preferentially selected during evolution is that they have default schema that equip them to visit many new cells on the grid rather than cycle over a few cells. The default schema of fit programs largely determines future optimization. The fit default schema are in most cases short and have a composition that leaves a program pointing in the same direction it was pointing at start of execution, so that subsequent program executions lead to new cell visits.

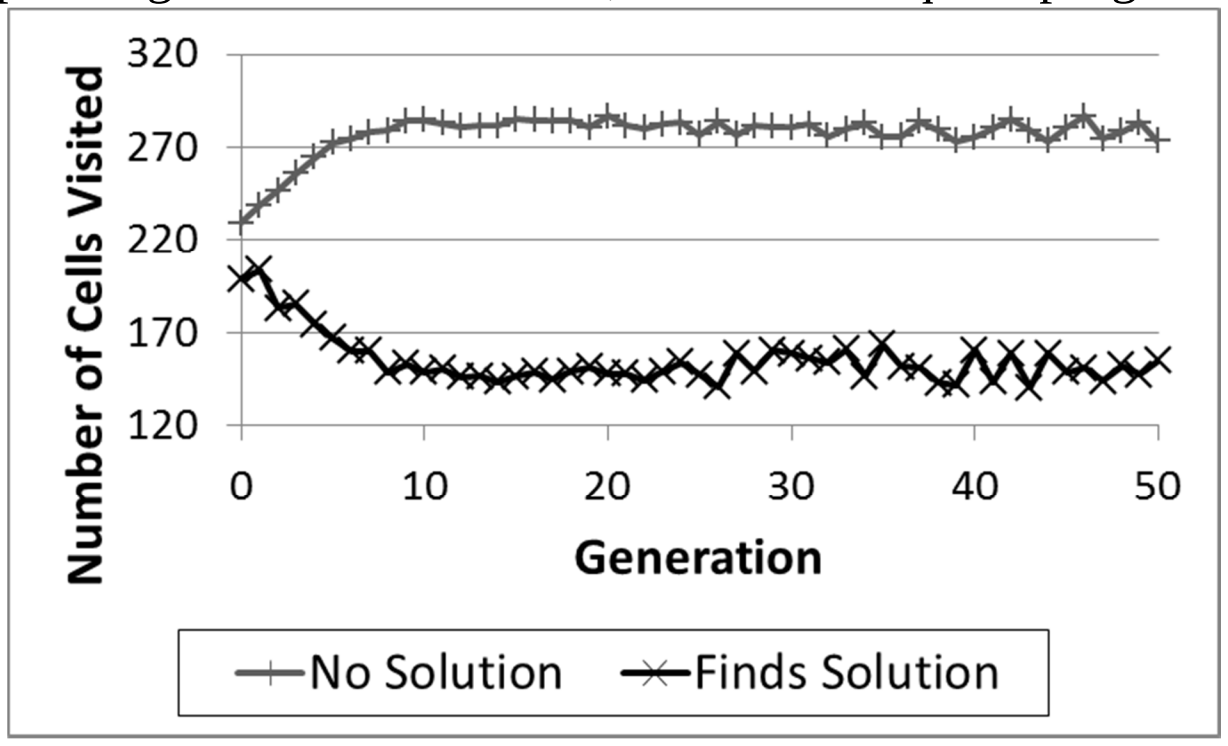

**Fig 19: Average number of different cells visited by fittest programs for evolutionary runs that find and don’t find solutions**

Langdon [11] pointed out that programs with zero and very low fitness are produced even in the final generation of evolutionary runs on this problem. This can be explained by the fact that the small changes (such as adding a single “left” instruction to a schema) modifies its direction and reduces the fitness of the program it is in. We also see that a schemas direction (and consequently its fitness) is determined by its whole composition. We cannot therefore associate fitness with a subsection of a schema, or attribute fitness improvement to an assembly process that puts together above average schema.

We divided the 6000 trials of experiment 3 into two sets: those that found a solution and those that did not. Fig 19 shows the evolution of the average number of unique cell visits for the fittest programs in both set. Fig 19 shows that generations that do not achieve solutions increase fitness by increasing the number of cells their programs visit. The average number of cells visited by trials that lead to solution drops and stabilizes close to a value of 144 (which is the number of cells on the trail).

Only about 13% of trials of standard GP on the Santa Fe Ant problem leads to solutions. Most programs with high fitness schema do not optimize towards following the trail; rather they achieve high fitness by visiting many cells independent of the trail structure. These programs that do not lead to solutions further optimize by developing more schema to respond to different trail conditions. Fig 20 shows the growth of the number of schema for runs that do not lead to solutions.

A small set of high fitness schema have the capacity to follow the food trail and further optimize to find solutions. In addition to the other properties that give default schemas high fitness, schemas that lead to solutions do a local search of immediate neighboring cells when they do not find food.

One reason why the proportion of programs that lead to solutions is small compared to those that do not follow the track is the synchronization requirements for following the track. Programs that follow the track need to "pass information" about the direction of the track to the next repetition of the program. In most cases this is done by programs ensuring that at the end of each iteration, they always points in the same direction (relative to the direction of the track) for all instruction sequences responding to any trail condition.

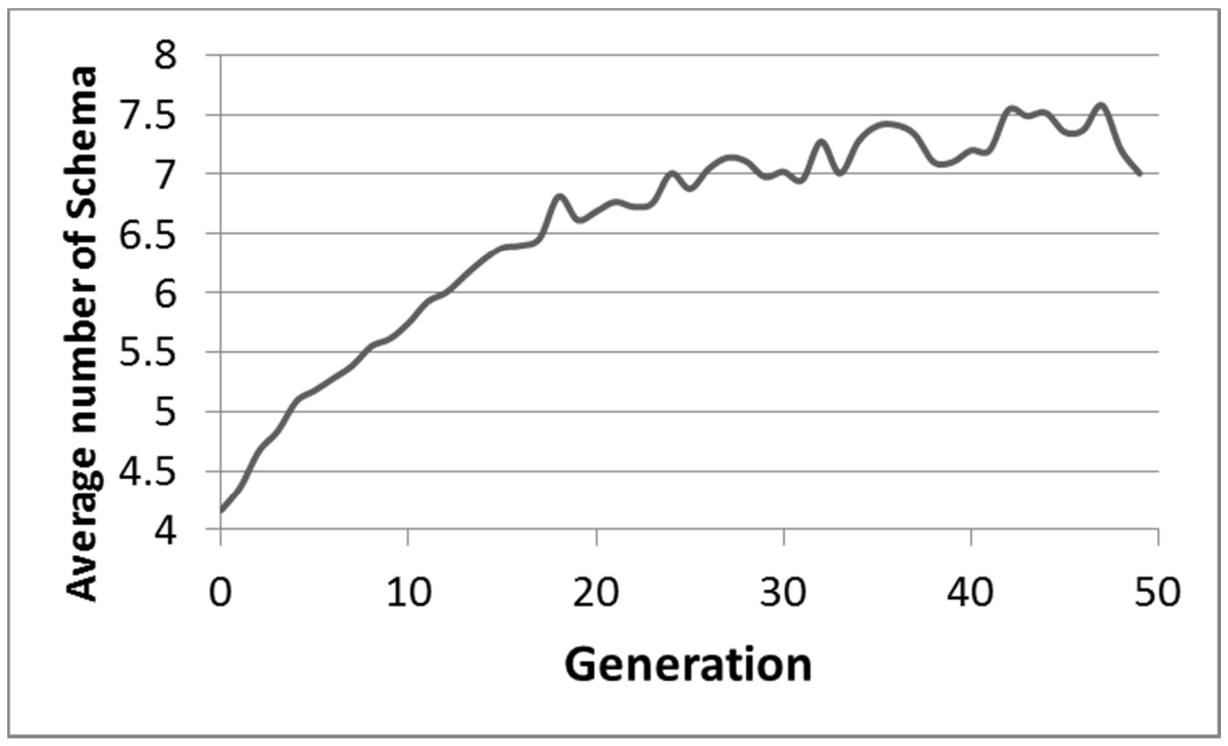

**Fig 20: Average number of schema by generation for runs that do not lead to solutions**

## VII. PERFORMANCE IMPROVEMENTS

We used the findings on the landscape of the default schema and the synchronization constraints to develop the following two methods of improving performance.

### *A. Phenotypic Crossover*

The first improvement method is to perform selective crossover that favors high fitness searches. We are aware that short, high fitness default schemas have a composition that leads to Eq (IV.1) evaluating to zero (i.e. they keep their initial direction). Our intention for the new phenotypic schema crossover is that it should conserve the attribute that is associated with fitness and thereby concentrate searching in high fitness schema areas.

Fig 21 shows the algorithm for our phenotypic crossover. Our crossover proceeds as the usual crossover operator, but only allows crossover for the case where the direction of the default schema of the subtree to be replaced is identical to the direction of the replacement subtree. When the subtrees do not have the same default schema directions we retry the crossover operation using different subtrees and crossover points until we achieve equal directions.

In the algorithm we do not just check for the directions of the subtrees being equal (i.e. we do not only use dir(Subtree (1)) = dir(Subtree (2)); This avoids a bias favoring the forward direction.

```
Choose k from 0 to 3 with equal probability ;
n = 0;
While (n<50)
    Select Subtree (1) from parent (1);
    Select Subtree (2) from parent (2);
    if (dir(Subtree (1)) = dir(Subtree (2)) = k)
        Crossover Subtrees;
        Break
    end if
    n = n + 1;
end While
```

**Fig 21: Algorithm for phenotypic crossover**

Note that this operation does not involve evaluating the trees or subtrees to be crossed over; it involves finding the composition of the default schema of the subtrees and computing their direction using Eq (IV.1).

### *B. New Representation*

In order to overcome the synchronization requirements faced by programs that optimize to the trail, we developed the new representation in Table 3 that reduces those requirements. An example of a solution using our new representation is shown in Fig 22.

In this new representation, we use an Automatically Defined Function (ADF). The Ifc-ADF ("If with continue" ADF) function is a unary function that calls the ADF function on finding food and operates as the usual IfFoodAhead function on not finding food. On finding food, a program will stop execution after executing the ADF and continue execution from the root of the program while retaining all state information. This is similar to how the "continue" statement is used in C-based languages.

The reasoning behind using Ifc-ADF is so that conditional functions in the main program can have the same instructions on finding food (due to executing the same ADF and no further code). This leads to synchronized responses on finding food and lower synchronization requirement for following the trail.

The ADF's contains the IfFoodAheadc binary function. This function operates as the IfFoodAhead function, however if it evaluates to true (i.e. food is found) the program continues from its root after executing its left branch as described for Ifc-ADF above.

**Table 3: New Representation**

| | **Functions and terminals** |
|---|---|
| Main Program | Ifc-ADF, ADF, PROG2, PROG3, LEFT, RIGHT, MOVE |
| ADF | IfFoodAheadc, PROG2, PROG3, LEFT, RIGHT, MOVE |

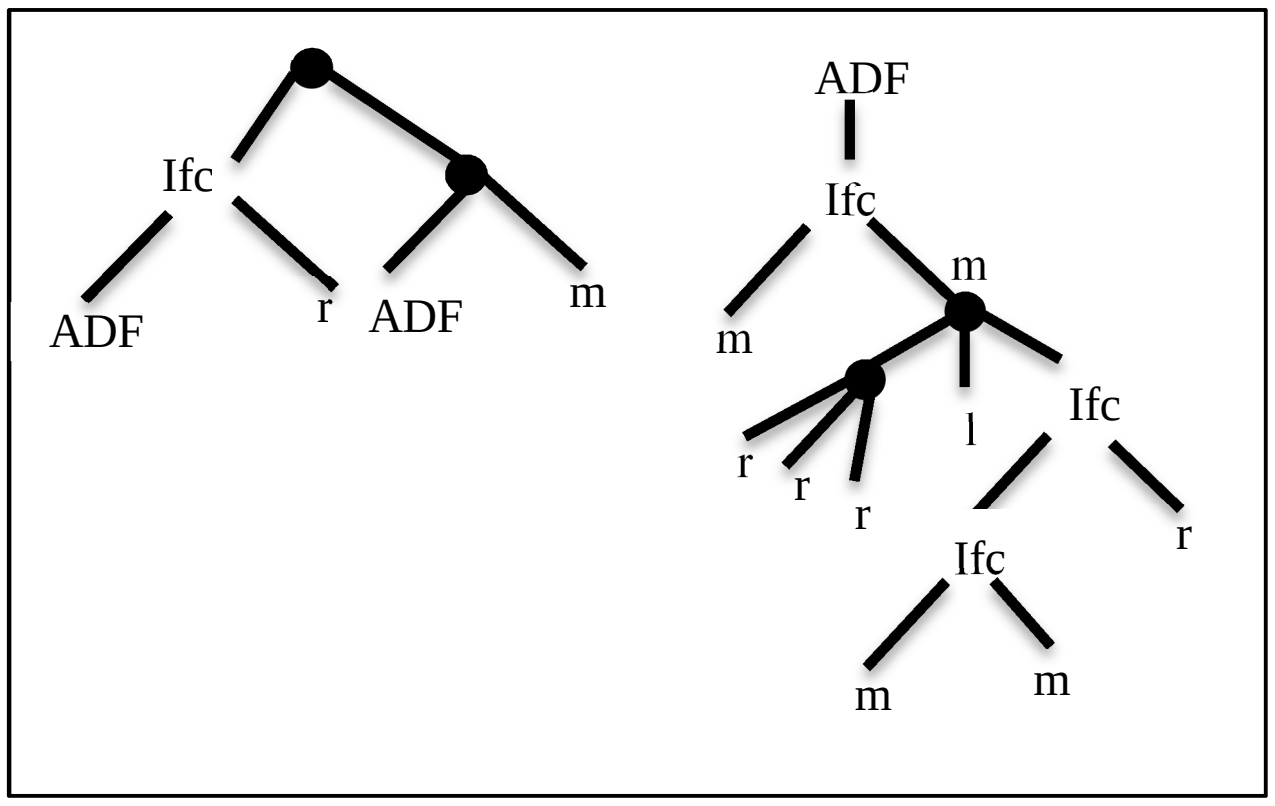

**Fig 22: Main program and ADF of example solution using new representation**

### *C. Results*

**Table 4: Performance Comparisons**

| **Santa Fe Ant Trail** | Success Rate | Computational Effort (x 1,000) |
|---|---|---|
| Standard GP [1,2] | 10.6% | 450 |
| GP with Phenotypic Crossover | 23.8% | 192 |
| GP with Alternative Representation | 72.7% | 67 |
| Both Phenotypic Crossover and Alternative Representation | **75.3%** | **51** |
| | | |
| **San Altos Trail** (population size 2000, Energy 3000, 100 generations) | | |
| Standard GP | 10.8% | 6,290 |
| GP with Phenotypic Crossover | 23.8% | 2,207 |
| GP with Alternative Representation | 96.4% | 85 |
| Both Phenotypic Crossover and Alternative Representation | **96.5%** | **75** |

The results in Table 4 show improvements in the success rate and in the computational effort required to generate solutions, when using either the phenotypic crossover or the alternative representation, or both. The improvements are also evident when applied to the more difficult San Altos trail.

## VIII. Conclusion

A poignant question for Evolutionary Computing practitioners is whether and when there is an expectation of a systematic structure between randomly generated programs and their fitness, and whether and how that structure is exploitable for more efficient search. For the Santa Fe ant problem we have discovered this structure; highly fit programs compose their executed instructions by using implicit repetition of short fit schema.

We have used executed instruction schema to analyze the relationship between program composition, behavior and fitness for the Santa Fe Ant problem. In doing this we discovered the schema compositions that are associated with high fitness and the geometry of the arrangement of such compositions. Since our findings on fitness structure and landscape are based only on analyzing random programs, they are general to the problem representation we use, irrespective of metaheuristic.

We ran evolutionary experiments specific to GP to monitor the dynamics of high fitness phenotypic schema. We then developed a description of how GP searched the landscape of this problem. We discovered two main approaches to optimization. The more prevalent approach optimizes by visiting more cells, consequently finding more food irrespective of the structure of the trail. The less prevalent approach leads to solutions by following the trail structure and optimizes by increasing its efficiency in following that structure.

We envisage further study on the use of executed instruction schema in discovering the intrinsic properties of programs to inform the choice of metaheuristic, variation operators and representation for nontrivial problems. In support of this approach we developed and tested a variation operator (based on the discovered geometry arrangement of schema fitness), and a representation method (designed to reduce discovered synchronization constraints). We recorded performance improvements for both changes on the Santa Fe and Los Altos trails. We plan to do research on autonomous applications of executed instruction schema to the discovery of the fitness structure in other programming problems, decision trees and rule based systems.

[50] Riccardo Poli. Exact schema theorem and effective fitness for GP with one-point crossover. In D. Whitley, D. Goldberg, E. Cantu-Paz, L. Spector, I. Parmee, and H.-G. Beyer, editors, Proceedings of the Genetic and Evolutionary Computation Conference, pages 469–476, Las Vegas, July 2000. Morgan Kaufmann.
[51] Riccardo Poli. Exact schema theory for genetic programming and variablelength genetic algorithms with one-point crossover. Genetic Programming and Evolvable Machines, 2(2):123–163, 2001.
[52] W. B. Langdon and Riccardo Poli. Foundations of Genetic Programming. Springer-Verlag, 2002.
[53] Riccardo Poli and Nicholas Freitag McPhee. General schema theory for genetic programming with subtree-swapping crossover: Part I. Evolutionary Computation, 11(1):53–66, March 2003.
[54] Riccardo Poli and Nicholas Freitag McPhee. General schema theory for genetic programming with subtree-swapping crossover: Part II. Evolutionary Computation, 11(2):169–206, June 2003.
[55] Riccardo Poli, Nicholas Freitag McPhee, and Jonathan E. Rowe. Exact schema theory and markov chain models for genetic programming and variable-length genetic algorithms with homologous crossover. Genetic Programming and Evolvable Machines, 5(1):31–70, March 2004.